\documentclass[sigconf,nonacm]{acmart}
\AtBeginDocument{%
  }

\setcopyright{acmlicensed}
\copyrightyear{2018}
\acmYear{2018}
\acmDOI{XXXXXXX.XXXXXXX}
\acmConference[Conference acronym 'XX]{Make sure to enter the correct
  conference title from your rights confirmation email}{June 03--05,
  2018}{Woodstock, NY}
\acmISBN{978-1-4503-XXXX-X/2018/06}

\usepackage{subcaption}
\usepackage{booktabs}
\usepackage{makecell}
\usepackage[table]{xcolor}
\usepackage{multirow} 
\usepackage[most]{tcolorbox}
\usepackage{xcolor}
\usepackage{algorithm}      
\usepackage{algpseudocode}

\usepackage{pgfplots}
\pgfplotsset{compat=1.18}

\usepackage{pifont}
\usepackage{enumitem}
\usepackage{tabularx}
\usepackage{booktabs}
\definecolor{MidOrange}{HTML}{E67E22}
\definecolor{HiGreen}{HTML}{1B9E77}
\newcommand{\midlvl}{\textcolor{MidOrange}{\textbf{Mid}}}
\newcommand{\highlvl}{\textcolor{HiGreen}{\textbf{High}}}

\newcommand{\cmark}{\textcolor{HiGreen}{\ding{51}}} 
\newcommand{\xmark}{\textcolor{red}{\ding{55}}}

\begin{document}

\title{WebGen-V Bench: Structured Representation for Enhancing Visual Design in LLM-based Web Generation and Evaluation}
\author{Kuang-Da Wang}
\authornote{Both authors contributed equally to this research.}
\authornote{Work done as an intern at Sony Group Corporation.}
\affiliation{\institution{National Yang Ming Chiao Tung University, Sony Group Corporation} \country{Tokyo, Japan}}

\author{Zhao Wang}
\authornotemark[1]
\authornote{Corresponding author: Zhao Wang (Email Address:
Zhao.Wang@sony.com)}
\affiliation{\institution{Sony Group Corporation} \country{Tokyo, Japan}}

\author{Yotaro Shimose}
\affiliation{\institution{Sony Group Corporation} \country{Tokyo, Japan}}

\author{Wei-Yao Wang}
\affiliation{\institution{Sony Group Corporation} \country{Tokyo, Japan}}

\author{Shingo Takamatsu}
\affiliation{\institution{Sony Group Corporation} \country{Tokyo, Japan}}


\renewcommand{\shortauthors}{Wang et al.}

\begin{abstract}
Witnessed by the recent advancements on leveraging LLM for coding and multimodal understanding, we present WebGen-V, a new benchmark and framework for instruction-to-HTML generation that enhances both data quality and evaluation granularity. WebGen-V contributes three key innovations: (1) an unbounded and extensible agentic crawling framework that continuously collects real-world webpages and can leveraged to augment existing benchmarks; (2) a structured, section-wise data representation that integrates metadata, localized UI screenshots, and JSON-formatted text and image assets, explicit alignment between content, layout, and visual components for detailed multimodal supervision; and (3) a section-level multimodal evaluation protocol aligning text, layout, and visuals for high-granularity assessment. Experiments with state-of-the-art LLMs and ablation studies validate the effectiveness of our structured data and section-wise evaluation, as well as the contribution of each component. To the best of our knowledge, WebGen-V is the first work to enable high-granularity agentic crawling and evaluation for instruction-to-HTML generation, providing a unified pipeline from real-world data acquisition and webpage generation to structured multimodal assessment.
\end{abstract}

\ccsdesc[500]{Computing methodologies~Natural language generation}
\ccsdesc[300]{Information systems~World Wide Web}
\ccsdesc[300]{Information systems~Evaluation of retrieval results}
\ccsdesc[100]{Computing methodologies~Artificial intelligence}

\keywords{Instruction-to-HTML Generation, 
Web Crawling, 
Structured Data Benchmark, 
Section-wise Evaluation, 
Multimodal Understanding
}


\begin{teaserfigure}
  \includegraphics[width=0.99\textwidth]{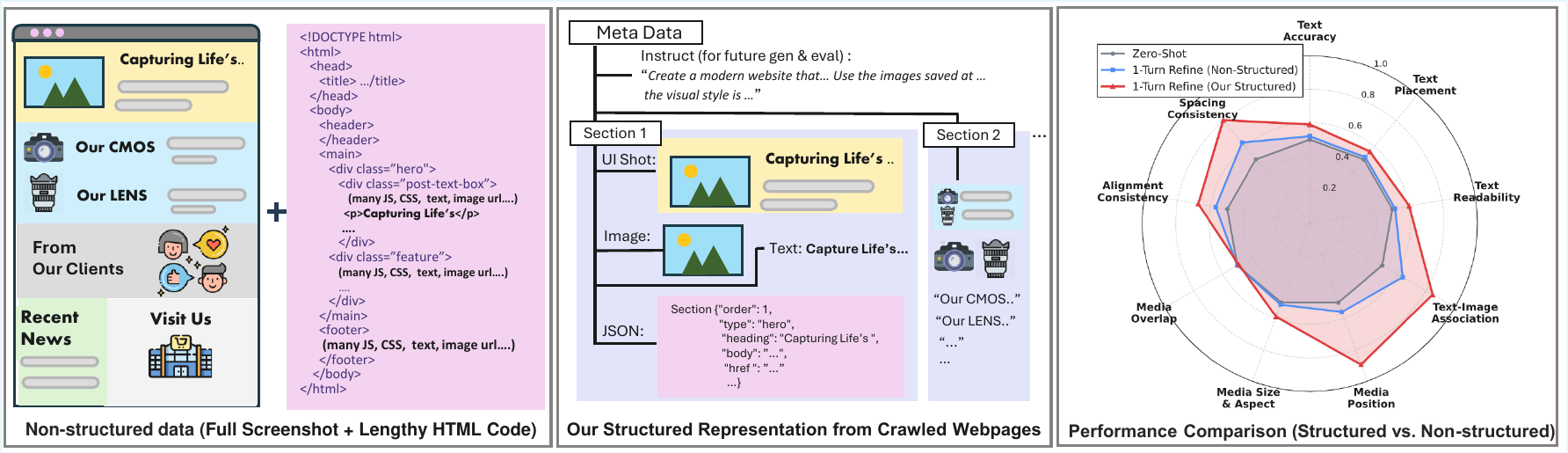}
  \caption{
    Comparison between traditional benchmarks and WebGen-V.
    \textbf{Left:} Existing benchmarks rely on full-page screenshots and lengthy raw HTML, which limit generation quality and fine-grained evaluation. 
    \textbf{Middle:} WebGen-V introduces structured, section-level data representations with localized UI screenshots, aligned text, image assets, and JSON metadata, enabling richer multimodal supervision and more precise evaluation. 
    \textbf{Right:} Compared to zero-shot generation or traditional 1-turn refinement using full-page screenshots and whole HTML, 1-turn refinement based on our structured data leads to higher-quality results. 
  }
  \label{fig:teaser}
\end{teaserfigure}


\maketitle

\section{Introduction}

\begin{table*}[t]
\centering
\small
\setlength{\tabcolsep}{4.5pt}
\caption{
    Comparison with related benchmarks. 
    \textbf{Data}: size, source, content type (\textit{HTML}: raw HTML file; \textit{ScrShot}: full screenshot; \textit{Meta}: metadata; \textit{SecShot}: per-section UI screenshots; \textit{JSON}: text extracted into structured JSON; \textit{Img}: extracted images), and average media count. 
    \textbf{Evaluation}: human consistency (Human), LLM refinement via feedback (Feed2LLM), section-wise textual evaluation (Sec-Text), and screenshots (ScrShot). Note that for WebMMU, only 38 webpages provide downloadable image links.
}
\begin{tabular}{@{}l cccc cccc@{}}
\toprule
\multirow{2}{*}{\textbf{Benchmark}} &
\multicolumn{4}{c}{\textbf{Data}} &
\multicolumn{4}{c}{\textbf{Evaluation}} \\
\cmidrule(lr){2-5}\cmidrule(lr){6-9}
& \textbf{Size} 
& \textbf{Source} 
& \textbf{Content} 
& \textbf{Avg. Media} 
& \textbf{Human} 
& \textbf{Feed2LLM} 
& \textbf{Sec-Text} 
& \textbf{ScrShot} \\
\midrule
WebBench~\cite{xu2025webbench} 
    & 1,000 & Human-Written & Instruct	 & 0 & \midlvl & \xmark & \xmark & \xmark \\
WebGen-Bench~\cite{lu2025webgenbench} 
    & 647 & Human \& GPT-4 Crafted & Instruct	 & 0 & \midlvl & \xmark & \xmark & \xmark \\
WebChoreArena~\cite{miyai2025webchorearena} 
    & 532 &  Curated Tasks & Instruct	 & - & \midlvl & \xmark & \xmark & \xmark \\
FullFront~\cite{sun2025fullfront} 
    & 1,800 & Model-Synthesized & HTML/ScrShot/Instruct & 1 & \midlvl & \xmark & \xmark & \textcolor{orange}{\textbf{LowQlty}} \\
WebDev Arena~\cite{lmsys2024webdevarena} 
    & N/A & Real User Prompts & HTML+JSON & - & \textbf{Optimal} & \xmark & \xmark & \xmark \\
WebMMU~\cite{awal2025webmmu} 
    & 2,059 (38) & Real Website+ScrShots & HTML/ScrShot/Instruct & 1 & \midlvl & \xmark & \xmark & \textcolor{orange}{\textbf{LowQlty}} \\
ArtifactsBench~\cite{xu2025artifactsbench} 
    & 1,825 & Expert \& LLM Curated & Instruct & - & \highlvl & \xmark & \xmark & \textcolor{orange}{\textbf{LowQlty}} \\
\midrule
\textbf{WebGen-V (Ours)} 
    & \makecell{\textbf{3,000}\\\textbf{+ Existing}} 
    & \textbf{Real-World Web Pages}
    & \makecell{\textbf{HTML/Meta/SecShot/}\\\textbf{JSON/Img/Instruct}} 
    & \textbf{23} 
    & \highlvl & \cmark & \cmark 
    & \textcolor{HiGreen}{\textbf{HighQlty}} \\
\bottomrule
\end{tabular}
\label{tab:related_work}
\end{table*}

Web page generation (WebGen) has emerged as a crucial research area at the intersection of natural language processing, multimodal learning, and software engineering. The web is the backbone of modern digital infrastructure~\cite{malecki2002internet,okelly2002backbone}, enabling commerce, education, media, and enterprise systems worldwide. The global web design market alone was valued at 52.01 billion USD in 2023 and is projected to grow to 108.38 billion USD by 2032~\cite{skyquest2024webdesign}, underscoring both the scale of industry demand and the need for automation. 

With the rapid progress of large language models (LLMs) in code generation~\cite{wang2024codeact,code_llm}, multi-modal understanding~\cite{wang2025mimo}, multi-agent cooperation~\cite{wang2025talkhier} and various applications~\cite{wang2025oms,wang2025okg,newsagent}, researchers have begun exploring web interface automation~\cite{Mind2Web,GPT-4V-Agent,multi_turn}, creation~\cite{pix2code,LayoutNUWA}, and evaluation~\cite{AgentAB}.
Early benchmarks approached web-related generation from different perspectives: 
\textit{WebBench}~\cite{xu2025webbench} focused on code generation involving standard web languages such as HTML and CSS through multi-step programming tasks, while \textit{WebGen-Bench}~\cite{lu2025webgenbench} targeted instruction-to-website generation with functional test-based evaluation, yet without paired visual references or screenshots.
\textit{FullFront}~\cite{sun2025fullfront} expanded this direction to multimodal front-end workflows, assessing LLMs on visual design comprehension, layout reasoning, and code synthesis. 
To capture visual and interactive fidelity, \textit{ArtifactsBench}~\cite{xu2025artifactsbench} introduced a framework for evaluating rendered webpage artifacts via multimodal comparison and LLM-as-judge scoring. 
Benchmarks like \textit{WebChoreArena}~\cite{miyai2025webchorearena} shifted focus toward web navigation and task execution, testing reasoning and memory in long-horizon agentic interactions rather than code generation. 
Similarly, the \textit{WebMMU}~\cite{awal2025webmmu} benchmark bridges webpage understanding and code synthesis, pairing a visual QA task for comprehension with synthesis tasks focused on code editing and mockup-to-code generation, rather than instruction-to-HTML.
Collectively, these works reveal a research trend from synthetic, static page-level evaluation toward multimodal and interaction-rich benchmarks for web intelligence.

Despite this steady progress, existing benchmarks remain insufficient for capturing the full complexity of real-world web generation tasks. As summarized in Table~\ref{tab:related_work}, most prior datasets suffer from several key limitations. 
First, many are either  synthetic~\cite{xu2025artifactsbench}, contain only generation instructions without paired HTML code~\cite{lu2025webgenbench,xu2025webbench,xu2025artifactsbench}, or lack visual richness~\cite{awal2025webmmu,lu2025webgenbench,xu2025artifactsbench}, providing few or no media elements such as images, and thus fail to reflect the realism and visual diversity of user-facing websites.
Second, although some benchmarks such as \textit{FullFront}~\cite{sun2025fullfront} and \textit{WebMMU}~\cite{awal2025webmmu} incorporate multimodal inputs, their data is still restricted to coarse full-page HTML and page-level screenshots, which makes fine-grained generation and alignment extremely difficult.
Moreover, most evaluation protocols~\cite{awal2025webmmu,xu2025artifactsbench} remain coarse, focusing primarily on overall correctness or human preference while failing to capture inconsistencies across sections, structural misalignment within layouts, or localized degradation in visual quality.
In summary, existing benchmarks provide only \textbf{oversimplified data} with \textbf{coarse-grained evaluation}, restricting models from learning high-quality, visually faithful webpage generation.

In this paper, we introduce \textbf{WebGen-V}, a multimodal web page benchmark designed to advance research in realistic and structured web page generation. 
Specifically, our originality lies in four key contributions:
\begin{itemize}
    \item \textbf{Unbounded and extensible real data construction.} We propose an \textit{agentic crawling framework} that is scalable and backward-compatible, which can continuously collect new real-world web pages and can be applied to augment existing benchmarks. 

    \item \textbf{Structured data representation.} Unlike prior benchmarks that only release raw HTML and page-level screenshots, WebGen-V provides structured outputs, including metadata, per-section UI screenshots, JSON-formatted text extractions, and extracted image/icon assets, that can be agentically and automatically generated. This structured design makes the benchmark not only richer but also more formalized for future large-scale evaluation. 

    \item \textbf{Section-wise multimodal evaluation.} By aligning structured data (text, images, UI shots, and metadata) at the section level, WebGen-V enables high-granularity evaluation that goes beyond whole-page judgments, allowing models to be assessed on localized content understanding and generation quality. 

    \item \textbf{Empirical demonstration.} We collect 3,000 new real-world webpages through our structured crawling pipeline, and show that the same framework can extend existing datasets. Experiments also confirm that section-wise evaluation and refinement yield consistent improvements in generation quality across models.

\end{itemize}

\begin{figure*}[t]
    \centering
    \includegraphics[width=0.9\textwidth]{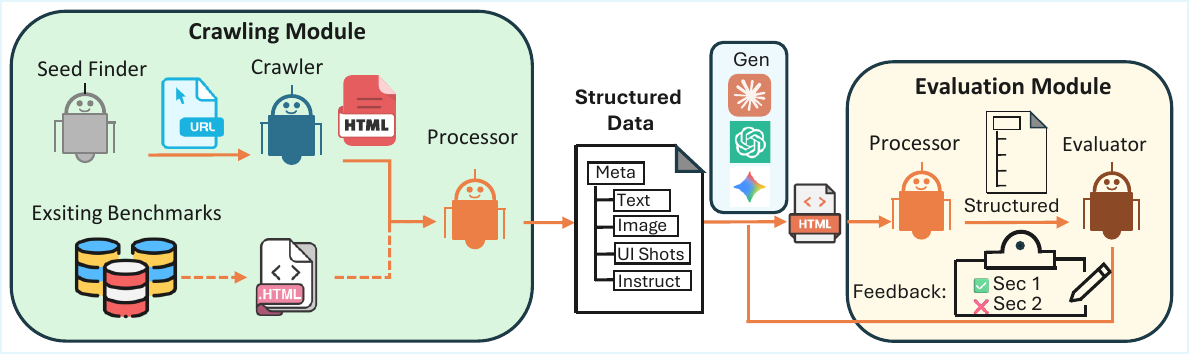}
    \caption{Overview of our framework consisting of two core modules, \textbf{Crawling} and \textbf{Evaluation}, which share a generic and scalable \textbf{Processor} for producing and analyzing structured webpage representations. The same Processor can also parse and transform existing benchmarks (e.g., \textit{WebMMU}), as illustrated by the dashed connections. Evaluation and feedback outputs can be seamlessly integrated with generation models (\textbf{Gen}) to enable iterative refinement and continuous improvement.}
    \label{fig:architecture}
\end{figure*}

\section{Related Work}
\label{sec:related_work}
Table~\ref{tab:related_work} summarizes existing benchmarks along two key dimensions: (\textbf{1})~the nature and richness of the data compositions, including realism and multimodal content; and (\textbf{2})~the scope and granularity of their evaluation protocols.

\paragraph{Code- and instruction-based generation.}
Early benchmarks such as \textit{WebGen-Bench}~\cite{lu2025webgenbench} and \textit{WebBench}~\cite{xu2025webbench} focus on code synthesis from textual instructions. Sourced from synthetic or real-world tasks, their complete absence of visual assets (e.g., images) constrains evaluation to code-level correctness, overlooking the critical challenge of multimodal layout design where text and visuals must be harmoniously integrated.

\paragraph{Web task automation.}
Benchmarks for web task automation, such as \textit{WebChoreArena}~\cite{miyai2025webchorearena}, test agentic reasoning and planning rather than generation. Consequently, they provide no supervision for the visual or structural fidelity of generated artifacts, which is the primary focus of our work.

\paragraph{Multimodal task.}
Recent benchmarks incorporating visual signals suffer from key limitations. \textit{FullFront}~\cite{sun2025fullfront} uses full-page screenshots, whose coarse representation and compression artifacts on long pages hinder fine-grained visual alignment. \textit{ArtifactsBench}~\cite{xu2025artifactsbench} advances evaluation by using temporal screenshots for dynamic interactions. However, its artifact-level judgments and limited images per task do not address the compositional challenge of generating long-form webpages, which requires coordinating numerous visual assets in a realistic layout.

\paragraph{Real-world web understanding.}
While sourced from real webpages, benchmarks for web understanding like \textit{WebMMU}~\cite{awal2025webmmu} are tailored for analytical tasks (e.g., VQA). Their reliance on coarse, full-page screenshots with few media elements does not provide the granular visual data needed to supervise high-fidelity, coordinate-aware layout generation.

\paragraph{Human preference evaluation.}
Platforms such as \textit{WebDev Arena}~\cite{lmsys2024webdevarena} provide a gold standard for assessing human preferences, yet they remain essentially ``black-box''.
Most current evaluations~\cite{xu2025artifactsbench} focus on coarse preference alignment with humans, leaving fine-grained feedback alignment unresolved.

\paragraph{Summary and Our Contributions.}
In summary, prior benchmarks are constrained by visually limited data and coarse, page-level representations. The lack of image assets hinders the evaluation of complex, coordinate-aware layout design—particularly on long pages where maintaining text–image harmony is essential. This creates a critical gap for benchmarks that can train and evaluate the generation of realistic, visually rich webpages at scale—a gap we address with WebGen-V.
\section{Methodology}
\label{sec:method}

Figure~\ref{fig:architecture} illustrates the overall \textbf{WebGen-V} framework, which provides a unified and structured pipeline for instruction-to-HTML benchmarking. The system centers on two primary components: a \textbf{Crawling Module} for large-scale acquisition and structured decomposition of real-world webpages, and an \textbf{Evaluation Module} for section-wise assessment of model outputs, both supported by a unified \textit{Processor} that converts raw HTML and visual content into fine-grained structured representations.

\subsection{Crawling Module}
\label{subsec:crawling}

The Crawling Module serves as the data acquisition and preprocessing backbone of WebGen-V. It continuously harvests webpages from diverse online domains and transforms them into structured data suitable for multimodal instruction generation and evaluation. The crawling result contains a wide variety of layouts, tones, and content types, covering corporate, product, portfolio, and education to ensure broad coverage of web design patterns. 
The complete list of keywords used for data collection is provided in Appendix~\ref{appendix:keywords}.

The process begins with a keyword-based \textit{Seed Finder}. A manually curated list of domain-relevant keywords (e.g., \texttt{crm}, \texttt{portfolio}, \texttt{booking}) is automatically expanded into high-intent queries such as “pricing,” “free trial,” or “contact us.” Each query is submitted to a search API~\cite{serpapi}, producing a ranked list of candidate URLs. After filtering for duplicates, invalid responses, and blacklisted domains (such as social media or video platforms), the crawler visits each remaining page using a hybrid renderer. Static pages are retrieved via lightweight HTTP requests, while JavaScript-dependent pages are rendered in full using Playwright\footnote{\url{https://playwright.dev/}} to ensure that dynamic Document Object Model (DOM) elements are captured. The crawler saves the full-page HTML, a rendered screenshot, and all embedded assets, including images, icons, and inline CSS. These raw outputs are then passed to the Processor for structured analysis. 

\subsubsection{Processor and Section-wise Structuring}
\label{subsec:processor}

The Processor is the central component that transforms raw webpages into multimodal structured data. Its purpose is to unify visual, textual, and structural modalities in a consistent and analyzable form. Rather than treating a webpage as a single monolithic entity, it performs \textit{section-wise decomposition} to identify distinct functional blocks—such as hero sections, product grids, testimonials, or footers—those represent coherent design and content units.
Given a page $W$, the Processor constructs:
\[
\mathcal{Z}(W) = \{S, T, I, M, B\},
\]
where $S$ denotes the ordered set of sections, $T$ the structured text content, $I$ the image assets linked to each section, $M$ the metadata including color palettes and typography, and $B$ the bounding boxes representing rendered coordinates in the visual layout.

The decomposition process (Algorithm~\ref{alg:sectionwise}) relies on both DOM-level parsing and visual heuristics:
\begin{enumerate}
  \item Identify candidate containers (\texttt{section}, \texttt{div}, \texttt{header}, \texttt{footer}) whose rendered height exceeds a threshold of 50 pixels, ensuring that only visually meaningful regions are considered.
  \item Merge visually overlapping containers when their bounding boxes intersect, ensuring that related elements (e.g., a headline and its corresponding CTA button) are grouped together without redundancy.
\end{enumerate}

Each resulting section $s_i$ is then independently rendered into a cropped screenshot to preserve its visual fidelity.
Associated text nodes, inline styles, and linked image references are extracted and stored as JSON entries.  
A LLM (\textit{GPT-5}) is then used to classify all detected image assets with semantic labels (i.e., “hero illustration,” “background image,” “icon,” “feature graphic,” and “logo”). These five categories are selected because they capture the most common functional roles of visual elements in modern web design.
This decomposition ensures that each webpage is represented as a balanced and interpretable set of sections, providing a foundation for both localized evaluation and multimodal supervision.

\begin{algorithm}[t]
\caption{\textsc{SectionWiseDecomposition}$(W)$}
\label{alg:sectionwise}
\small
\begin{algorithmic}[1]
\Require Webpage $W$ with HTML $H$ and rendered image $J$
\Ensure Section-wise representation $\mathcal{Z} = \{S, T, I, M, B\}$,  
where $S$: detected sections, $T$: text content, $I$: image assets, $M$: style metadata (colors, fonts), $B$: bounding boxes.
\State Parse $H$ into DOM tree $\mathcal{D}$
\State Identify top-level containers $\mathcal{C} \leftarrow \{\texttt{section}, \texttt{div}, \texttt{header}, \texttt{footer}\}$
\State Initialize section list $S \leftarrow \emptyset$
\For{each $c \in \mathcal{C}$}
  \State Compute bounding box $b_c$ under fixed viewport
  \If{$\text{height}(b_c) > \theta_{\min}$} \Comment{$\theta_{\min}$: min height}
    \State Append candidate section $(c, b_c)$ to $S$
  \EndIf 
\EndFor
\State Merge overlapping or adjacent sections if $\mathrm{IoU}(b_i,b_j) > 0$ 
\State Extract textual ($T$), visual ($I$), and style ($M$) metadata from finalized $S$
\State \Return $\mathcal{Z} = \{S, T, I, M, B\}$
\end{algorithmic}
\end{algorithm}

\subsubsection{Instruction Generation}
Using the structured output from the Processor, \textit{GPT-5} generates concise design specifications that capture each webpage’s purpose and visual intent.
Instead of reproducing the entire content, the instruction is designed to extract only the essential textual and visual elements required for reconstruction, while omitting layout templates and low-level stylistic details.
This shifts the emphasis from how the page is rendered to what it communicates.
The prompt takes as input the HTML, section-wise structured text, and image metadata classified by the vision-language model, summarizing them into a grounded, human-readable specification that includes only assets explicitly referenced in the HTML.

Specifically, each generated instruction includes:
\begin{itemize}
\item A concise overview describing the page’s design target;
\item Section-wise summaries derived from semantic compression, each linked to its corresponding cropped image and text segment, with layout details removed;
\item A list of visual assets with filenames and dimensions;
\item The primary calls-to-action and hyperlink destinations;
\end{itemize}
This distilled specification serves as a multimodal instruction that captures the essence of a webpage while preserving design flexibility for downstream HTML generation.

\subsection{Evaluation Module}
\label{subsec:evaluation}

The Evaluation Module assesses generated webpages through a structured, section-wise lens.
Crucially, this process does \textbf{not} rely on direct comparison with a reference layout or ground truth, since webpage design inherently allows multiple valid realizations for the same instruction—and because layout constraints are intentionally removed—there is no single “correct” solution.  
Instead, evaluation is conducted entirely by a multimodal LLM capable of reasoning over both textual and visual inputs. Following recent findings that GPT-5 achieves superior visual reasoning~\cite{wang2025multimodal, datacamp2025gpt5} and multimodal alignment~\cite{hu2025radiology}, we adopt GPT-5 as the evaluator backbone. 

\subsubsection{Evaluation Pipeline}
Given an instruction $\Sigma$ and the model-generated HTML $\hat{H}$, the evaluation proceeds as follows:
\begin{enumerate}
  \item Render $\hat{H}$ and apply the same Processor for section decomposition, obtaining $\hat{\mathcal{Z}} = \{\hat{S}, \hat{T}, \hat{I}, \hat{M}, \hat{B}\}$.
  \item For each section $\hat{s}_i$, the LLM evaluates text correctness, visual alignment, readability, and multimodal coherence.
  \item Instead of comparing against a gold reference, the LLM performs rule-based and contextual reasoning derived from the instruction $\Sigma$ to assess whether the design logically fulfills the described intent.
  \item The results are aggregated into structured feedback $\mathcal{F}$, where each entry contains a quantitative score and qualitative rationale.
\end{enumerate}

Formally, the evaluation produces:
\[
\mathcal{F} = \{(\hat{s}_i, m_k, \text{score}, \text{reason}, \text{feedback})\},
\]
where each tuple corresponds to a section $\hat{s}_i$, a metric category $m_k$ (e.g., readability, balance, alignment), and an actionable improvement feedback.

\subsubsection{Generation–Evaluation–Refinement Process}
The feedback $\mathcal{F}$ can optionally be used to refine the generated HTML.  
When any section’s score falls below a threshold $\tau$, the corresponding feedback is reintroduced into the model, prompting a targeted regeneration:
\[
\hat{H}' = f_{\text{refine}}(\Sigma, \hat{H}, \mathcal{F}).
\]
The refinement is guided directly by section-level feedback, allowing the model to address specific issues identified in each region, such as misaligned text blocks, unbalanced spacing, or missing media elements.  
This refinement preserves overall design flexibility while applying localized improvements based on explicit evaluations.
Overall, this module forms an integrated process: the generation step proposes, the evaluation step critiques, and the refinement step, if triggered, improves the design.  

\begin{algorithm}[t]
\caption{\textsc{Gen-Eval-Refine}$(\Sigma, \tau)$}
\label{alg:eval-refine}
\small
\begin{algorithmic}[1]
\Require Instruction $\Sigma$, refinement threshold $\tau$
\Ensure Final HTML $\hat{H}'$, feedback $\mathcal{F}$
\State Generate HTML: $\hat{H} = f_{\text{gen}}(\Sigma)$
\State Decompose sections: $\hat{\mathcal{Z}} \leftarrow \textsc{SectionWiseDecomposition}(\hat{H})$
\State Evaluate via LLM: $\mathcal{F} = f_{\text{eval}}(\hat{\mathcal{Z}}, \Sigma)$
\If{$\min(\text{score}) < \tau$}
  \State Refine HTML: $\hat{H}' = f_{\text{refine}}(\Sigma, \hat{H}, \mathcal{F})$
\Else
  \State $\hat{H}' = \hat{H}$ \Comment{No refinement needed}
\EndIf
\State \Return $\hat{H}'$
\end{algorithmic}
\end{algorithm}

\begin{figure*}[t]
    \centering
    \begin{subfigure}{0.32\textwidth}
        \includegraphics[width=\linewidth]{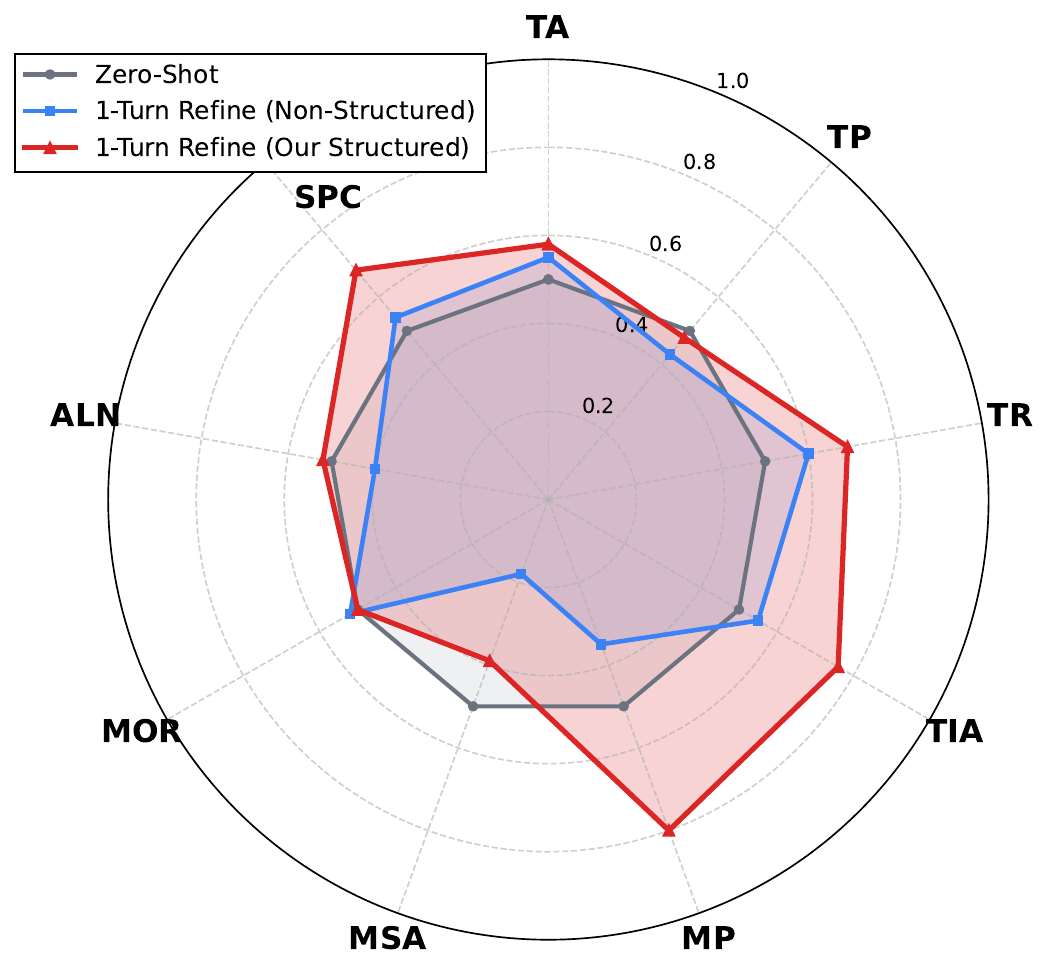}
        \caption{Claude-Opus-4.1}
    \end{subfigure}
    \begin{subfigure}{0.32\textwidth}
        \includegraphics[width=\linewidth]{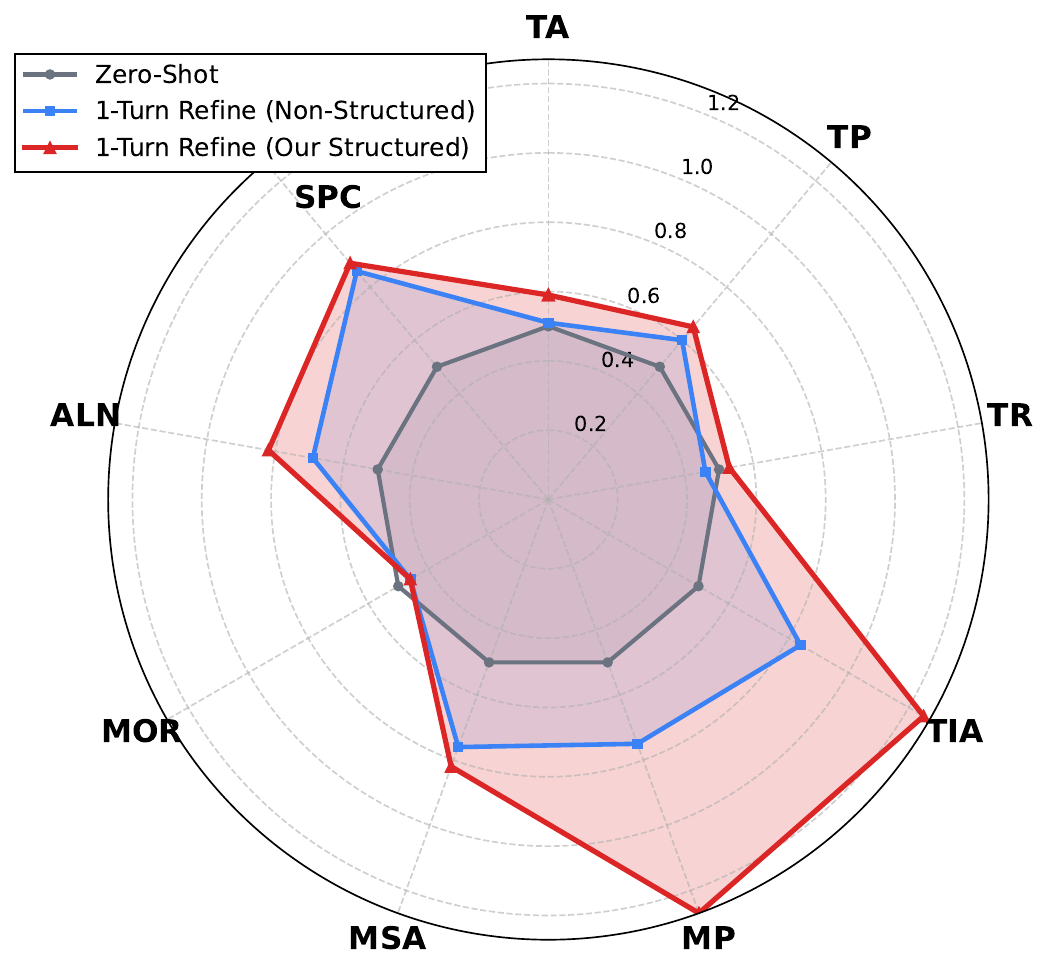}
        \caption{GPT-5}
    \end{subfigure}
    \begin{subfigure}{0.32\textwidth}
        \includegraphics[width=\linewidth]{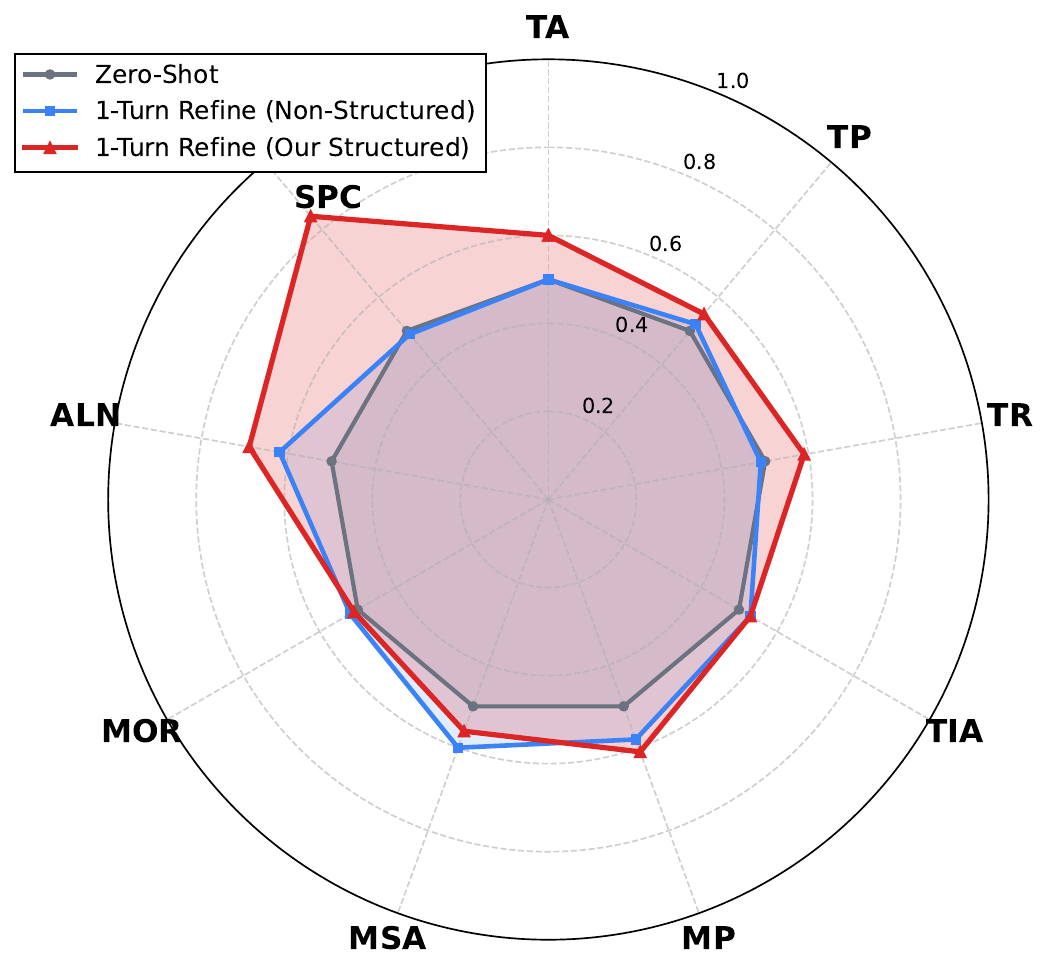}
        \caption{Gemini-2.5-Pro}
    \end{subfigure}
    \caption{Relative improvements over zero-shot generation across nine metrics for three backbone models: (a) Claude-Opus-4.1, (b) GPT-5, and (c) Gemini-2.5-Pro. Each radar plot shows the normalized gain achieved by one-turn refinement using either non-structured full-page input (blue) or our structured representation (red). Larger areas indicate stronger improvements.}
    \label{fig:radar-diff}
\end{figure*}
\vspace{4pt}

\begin{figure*}[h]
  \centering
  \begin{subfigure}[b]{0.48\textwidth}
    \centering
    \includegraphics[width=\textwidth]{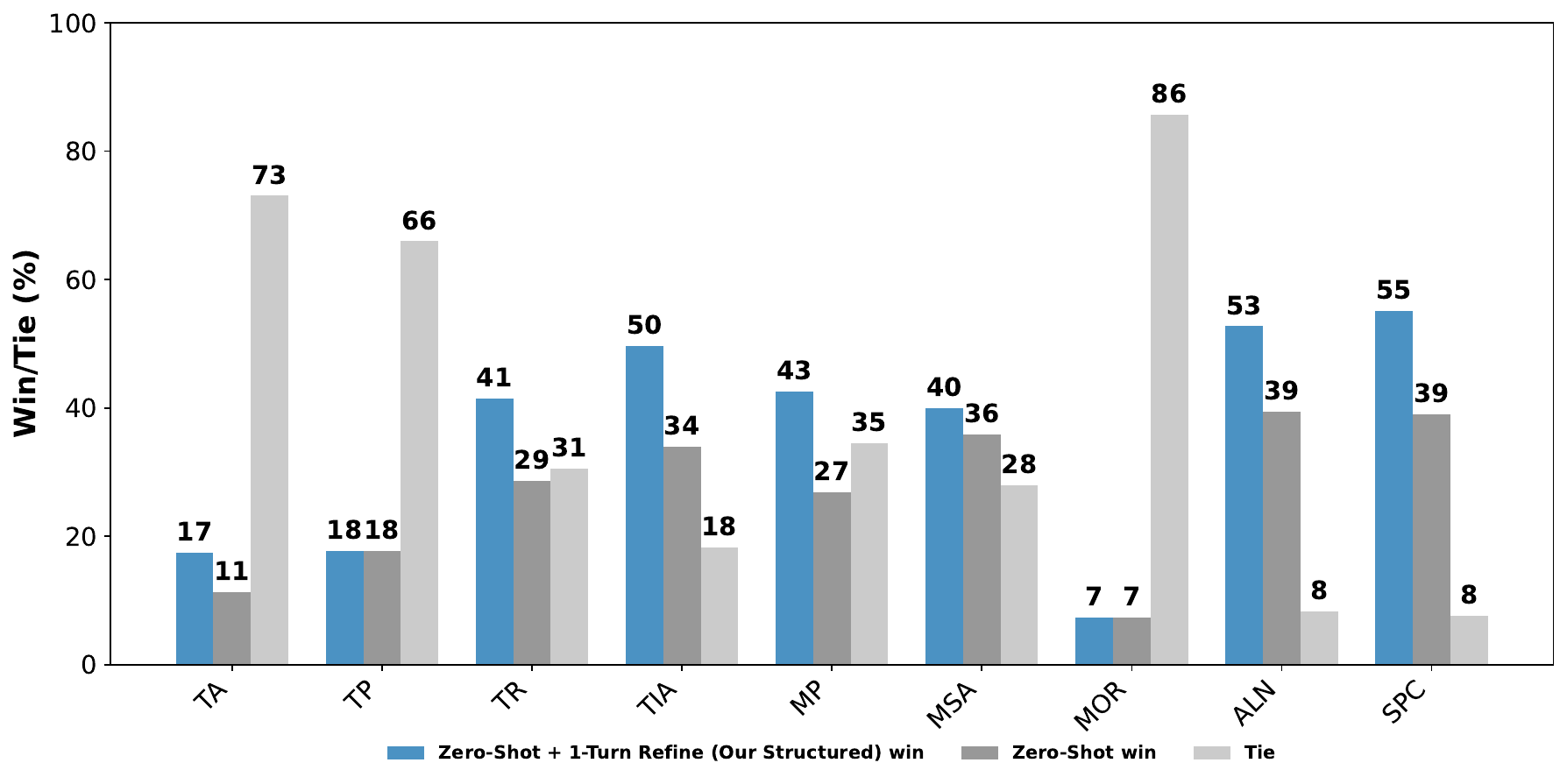}
    \caption{Zero-shot + 1-Turn Refine (Our Structured) vs.\ Zero-shot}
    \label{fig:avg_zero_vs_struct}
  \end{subfigure}
  \hfill
  \begin{subfigure}[b]{0.48\textwidth}
    \centering
    \includegraphics[width=\textwidth]{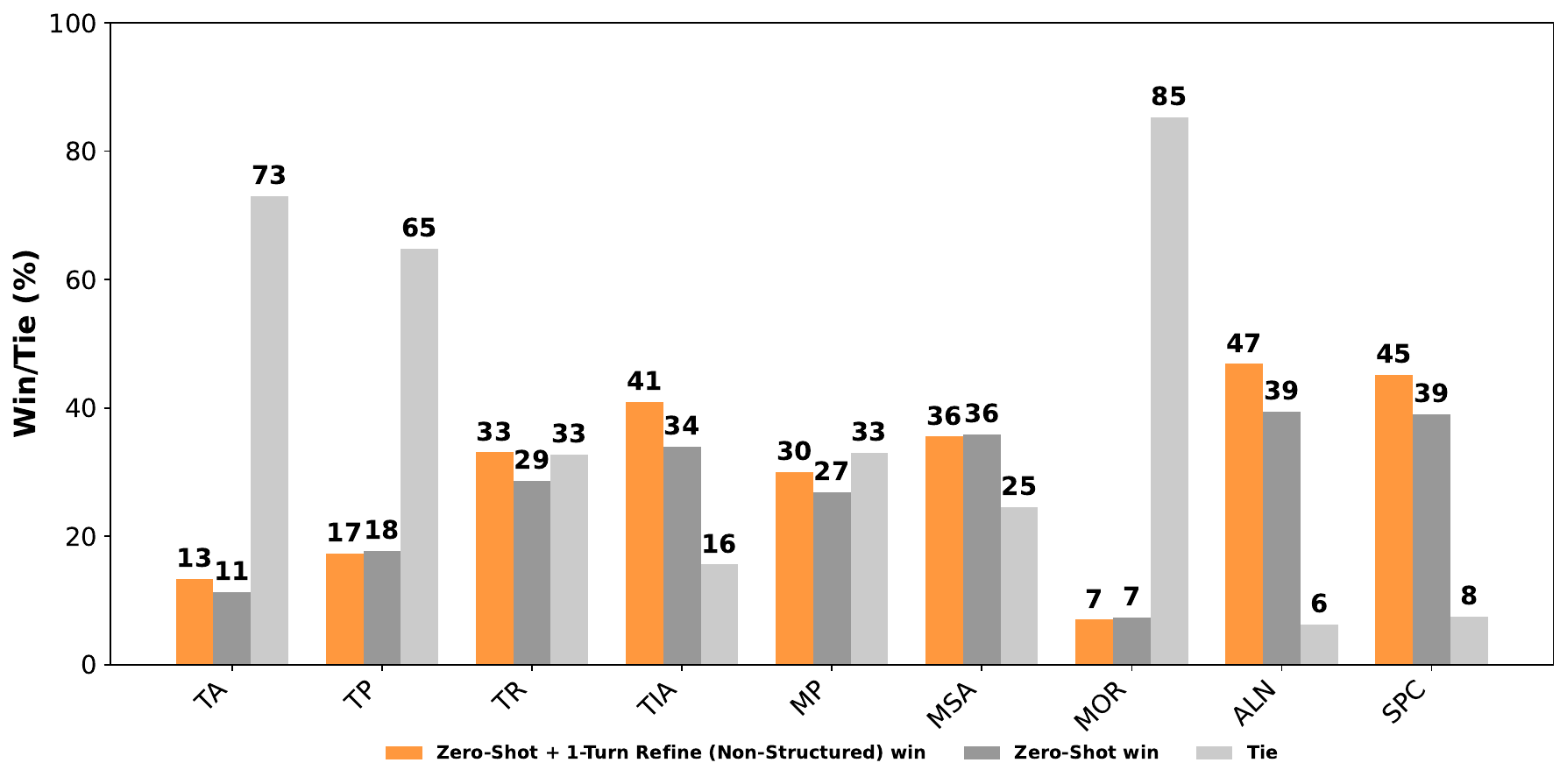}
    \caption{Zero-shot + 1-Turn Refine (Non-Structured) vs.\ Zero-shot}
    \label{fig:avg_zero_vs_whole}
  \end{subfigure}
  \caption{
    Head-to-head win-rate comparison averaged across all backbone models (\textbf{Claude-Opus-4.1}, \textbf{GPT-5}, and \textbf{Gemini-2.5-Pro}). Each bar represents the mean proportion of wins, ties, and losses under pairwise evaluation across all model backbones.
  }
  \label{fig:avg_pairwise_comparison}
\end{figure*}

\section{Experiments}
\label{sec:experiments}

We aim at answering the following research questions:

\begin{description}[leftmargin=1.5em,labelsep=0.5em]
\item[RQ1] \textbf{Does structured, section-wise representation enhance evaluation quality compared to full-page inputs?}  
We analyze whether fine-grained, structured inputs enhance evaluation accuracy and, in turn, provide more reliable feedback that contributes to improved generation quality across three state-of-the-art LLM backbones. (Section~\ref{sec:backbone-comparison}).
\item[RQ2] \textbf{Can the proposed structured evaluation module more accurately detect human-injected degradations than traditional full-page evaluation?}  
We design a controlled benchmark with recorded ground-truth degradations across text, layout, and media categories, and compare detection results between the traditional full-page evaluation and our structured, section-wise evaluation module.  (Section~\ref{sec:human_consistency}).
\item[RQ3] \textbf{Which components of structured data contribute most to evaluation quality?}  
We perform an ablation comparing section-wise screenshots only, full-page screenshots only, and full multimodal structured inputs (Section~\ref{sec:ablation}).
\item[RQ4] \textbf{Can our transformation module adapt existing HTML benchmarks into a structured format?}  
We conduct a case study on a web code editing task from WebMMU to evaluate the generality and scalability of our framework for transforming prior benchmarks (Section~\ref{sec:html-benchmark}).
\end{description}

\subsection{Experimental Setup}
\label{sec:data}

\paragraph{Data Crawling.} We employed our agentic crawling framework to collect a diverse set of real-world web page from multiple domains.
To ensure broad coverage and multifaceted applicability, we constructed a comprehensive keyword set comprising  \textbf{300} domain-specific terms spanning six major sectors: \textit{SaaS} (e.g., CRM, project management, marketing automation), \textit{Finance and Insurance} (e.g., credit cards, loans, trading platforms, insurance products), \textit{Health and Wellness} (e.g., telemedicine, fitness, nutrition, skincare), \textit{E-commerce} (e.g., fashion, electronics, home appliances, subscriptions), \textit{Travel and Lifestyle} (e.g., flight and hotel booking, tours, travel insurance), and \textit{Education and Professional Services} (e.g., online degrees, bootcamps, certification courses). 
Guided by these keywords, our crawler retrieved approximately \textbf{3,000} representative web pages across industries, reflecting realistic commercial web designs. 

\paragraph{Evaluation Metrics.} \label{sec:eval_metrics}
To evaluate instruction-based HTML generation, we employ a focused set of nine metrics across three categories. 
For \textbf{Text}, we measure \textit{Text Accuracy (TA)} for content correctness, \textit{Text Placement (TP)} for verifying that text appears in the intended section, and \textit{Text Readability (TR)} for ensuring sufficient visual contrast. 
For \textbf{Media}, we assess \textit{Text–Media Association (TIA)} to check semantic alignment between Media and captions, \textit{Media Positional Accuracy (MP)} for verifying Media placement within the specified container, and \textit{Media Size \& Aspect Compliance (MSA)} to confirm adherence to expected proportions. 
For \textbf{Layout}, we include \textit{Media Overlap Robustness (MOR)} to detect unintended collisions between images and other elements, \textit{Alignment Consistency (ALN)} to evaluate column and edge alignment, and \textit{Spacing Consistency (SPC)} to measure uniformity of vertical and horizontal gaps. 
Together, these metrics capture textual fidelity, media correctness, and structural layout quality at a fine granularity.

\paragraph{Scoring Protocol.}  
Each metric is rated on a 1–5 discrete scale, where higher values correspond to better quality. Threshold rules ensure interpretability (e.g., any section with visible errors cannot exceed a score of 3). 

To more comprehensively capture differences across systems, we employ two complementary comparison strategies:
(1)~\textbf{Error reduction analysis.}
Rather than comparing absolute scores, we quantify the improvement of each method relative to its zero-shot baseline.
For every metric, we measure the reduction in the number of low-scoring sections ($\leq 4$) after refinement, which highlights the model’s ability to correct local issues rather than merely increasing overall averages. 
It also reflects practical webpage quality, where even minor visual defects such as inconsistent spacing or misaligned elements can significantly affect overall design coherence.
(2)~\textbf{Head-to-head comparison.} To examine refinement performance, we directly compare paired outputs from the same input instruction, using the zero-shot result as the baseline. This dimension reveals whether refinements systematically correct existing issues instead of relying on potentially misleading aggregate statistics.

\paragraph{Generation and Evaluation Backbones.}  
For instruction-to-HTML generation, we evaluate three state-of-the-art closed-source multimodal LLMs: \textit{GPT-5}~\cite{openai2025gpt5}, \textit{Gemini-2.5-Pro}~\cite{google2025gemini25}, and \textit{Claude-Opus-4.1}~\cite{anthropic2025claude41}.  
All models are prompted with identical structured instructions specifying must-have content, layout constraints, and media references for fair comparisons.
The full prompt templates used throughout the study are provided in the Appendix: the instruction-generation prompt in Appendix~\ref{app:instruction_prompt}, the zero-shot generation prompt in Appendix~\ref{app:generation_prompt}, the evaluation prompt in Appendix~\ref{app:evaluation_prompt}, and the refinement prompt in Appendix~\ref{app:refinement-prompt}.  
A detailed comparison of the token efficiency of these models is provided in Appendix~\ref{sec:token_cost}.

\subsection{RQ1: Effectiveness of the Structured Crawling and Evaluation Module}
\label{sec:backbone-comparison}
Figure~\ref{fig:radar-diff} presents the relative improvements of refinements over zero-shot generation in terms of our proposed structured and section-wise feedback as well as non-structured refinement.
GPT-5 benefits the most, showing clear gains in Spacing Consistency (SPC), Text-Media Association (TIA), and Media Positional Accuracy (MP), where localized structural analysis reveals errors that are often invisible at the whole-page level.
Additionally, our structured refinement on Claude-Opus-4.1 and Gemini-2.5-Pro also yields noticeable gains in SPC and MP. In some cases, non-structured refinement even leads to degraded quality.
Win-rate comparisons in Figure~\ref{fig:avg_pairwise_comparison} further confirm that structured refinement produces more consistent improvements across samples, achieving over 50\% win rates in several layout-related dimensions such as SPC and ALN. 
Complete numerical results and the win rate comparisons among the three models are provided in Appendix~\ref{app:add_table}.

It is noteworthy that the errors produced by state-of-the-art models in zero-shot generation are rarely global structural flaws. Instead, they typically manifest as localized imperfections, such as inconsistent spacing or suboptimal color contrast. While these fine-grained issues are often imperceptible or overlooked in a holistic, full-page screenshot, they are readily apparent from a professional design perspective. 
The efficacy of our structured method stems from its ability to isolate and magnify these specific regions, making subtle errors identifiable and therefore reducible. In contrast, evaluations relying solely on full-page screenshots and whole HTML code tend to miss such localized inconsistencies, resulting in incomplete feedback and less effective refinement. We provide qualitative examples illustrating this difference in Appendix~\ref{app:sony_case}, and empirically validate this observation in the following subsection through quantitative comparisons between non-structured and structured settings.

\begin{tcolorbox}[colback=gray!5,colframe=gray!40,boxrule=0.4pt,arc=2pt,left=6pt,right=6pt,top=3pt,bottom=3pt]
\textbf{Takeaway.} Structured representation provides a more reliable and interpretable evaluation framework than whole-page inputs, consistently improving multimodal generation quality across models and metrics.
\end{tcolorbox}

\subsection{RQ2: Consistency with Human Judgment}
\label{sec:human_consistency}
Reliable refinement depends on accurately detecting genuine design errors.
Modern LLMs can effectively correct issues once they are explicitly localized and described.
We therefore evaluate whether our structured evaluation captures real degradations.

\textbf{Task formulation.}
To verify whether our proposed structured evaluation module can more accurately detect real degradations than conventional full-page evaluation, we design a controlled experiment with ground-truth annotations.  
We intentionally inject degradations into webpages across three categories:
(1) \textbf{Text}--font size, color, or readability changes;
(2) \textbf{Layout}--spacing, alignment, or positioning distortions;
(3) \textbf{Media}--image/video resizing or aspect-ratio alterations.  

Crucially, since we record the exact locations and types of all human-injected degradations, this dataset provides \textbf{precise ground-truth labels}.  
Thus, we can objectively measure whether each evaluation method (traditional or ours) correctly identifies the degraded sections.
We compare two evaluation settings:
\textbf{Non-Structured Evaluation} uses raw HTML with a full-page screenshot as input, typical of prior works.  
\textbf{Structured Evaluation (Ours)}: uses our section-wise structured data, including localized UI screenshots, extracted text, and metadata.




Each module outputs binary predictions (\emph{degraded} or \emph{clean}) for every section–category pair. Because both evaluation and feedback in our framework fundamentally depend on identifying the presence of errors, verifying that degradations are correctly detected is essential. In practice, once the evaluator successfully localizes a real error, LLMs can effectively correct it when this feedback is provided. We compute \textit{Precision}, \textit{Recall}, and \textit{F1-score}, reporting per-category results and both micro and macro averages.

\begin{table}[t]
\centering
\caption{Comparison of evaluation settings in detecting human-injected degradations. Our structured setting matches ground truth far more accurately than the non-structured one.}
\label{tab:structured-vs-full}
\begin{tabular}{lcccccc}
\toprule
& \multicolumn{3}{c}{\textbf{Non-Structured}} & \multicolumn{3}{c}{\textbf{Structured (Ours)}} \\
\cmidrule(lr){2-4}\cmidrule(lr){5-7}
\textbf{Category} & \textit{P} & \textit{R}& \textit{F1} & \textit{P} & \textit{R} & \textit{F1} \\
\midrule
Text   & 0.36 & 0.44 & 0.40 & \textbf{0.69} & \textbf{0.71} & \textbf{0.70} \\
Layout & 0.58 & 0.61 & 0.59 & \textbf{0.90} & \textbf{0.89} & \textbf{0.90} \\
Media  & 0.32 & 0.36 & 0.34 & \textbf{0.64} & \textbf{0.67} & \textbf{0.65} \\
\midrule
Micro Avg & 0.45 & 0.48 & 0.46 & \textbf{0.78} & \textbf{0.79} & \textbf{0.78} \\
Macro Avg & 0.42 & 0.47 & 0.44 & \textbf{0.74} & \textbf{0.76} & \textbf{0.75} \\
\bottomrule
\end{tabular}
\end{table}

Structured evaluation achieves substantially higher alignment with human-labeled degradations compared to full-page baselines, demonstrating its superior ability to localize and identify realistic layout, text, and media defects.
As shown in Table~\ref{tab:structured-vs-full}, our method improves average F1 from 0.46 to 0.78. This result confirms that section-wise representations enable evaluators to not only detect degradations but also pinpoint their locations accurately.
\begin{tcolorbox}[colback=gray!5,colframe=gray!40,boxrule=0.4pt,arc=2pt,left=6pt,right=6pt,top=3pt,bottom=3pt]
\textbf{Takeaway.} Section-wise structured evaluation provides a faithful proxy for human judgment, capturing degradations with higher precision and offering localized feedback that full-page non-structured evaluation cannot.
\end{tcolorbox}

\begin{figure}[t]
    \centering
    \includegraphics[width=\linewidth]{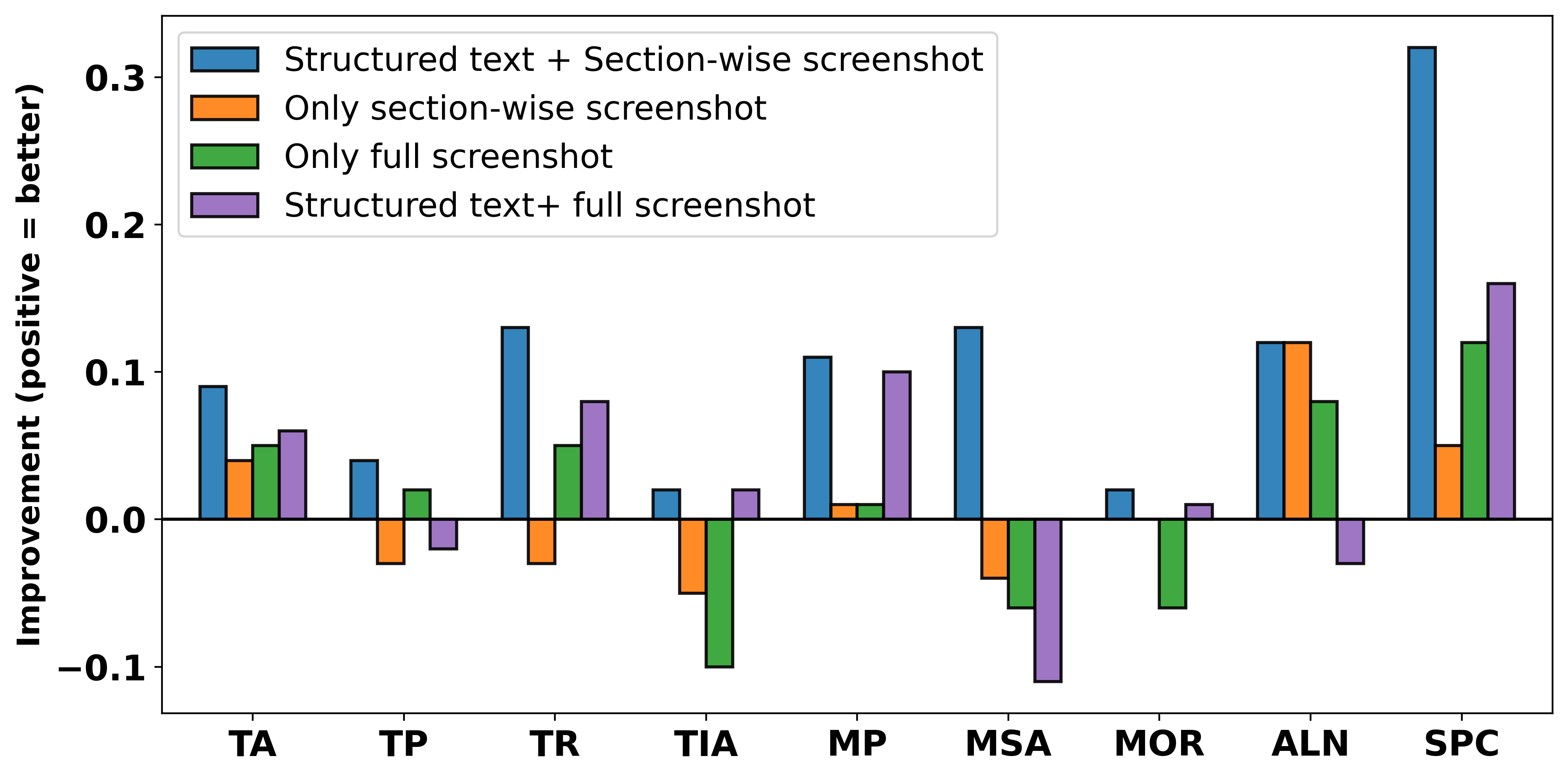}
    \caption{
        \textbf{Ablation study on different input modalities.} 
        Structured text and section-wise screenshots yield the strongest improvements over the \emph{Gemini-2.5 zero-shot} generation, particularly for readability, spacing, and media placement. Full-page inputs blur fine-grained errors and often regress.
    }
    \label{fig:ablation}
\end{figure}

\subsection{RQ3: Ablation Analysis} \label{sec:ablation}
As presented in Figure~\ref{fig:ablation}, we assess the importance of different input modalities by analyzing combinations of structured text, section-wise screenshots, and full-page screenshots. The results reveal a clear pattern: \textbf{fine-grained, section-wise cues are essential}.  

Combining structured text with section-wise screenshots yields the most significant performance gains across all metrics, particularly for \textit{Text Readability (TR)}, \textit{Media Positional Accuracy (MP)}, and \textit{Spacing Consistency (SPC)}. These metrics are sensitive to localized features, such as font contrast or intra-container image alignment, which are effectively captured by section-wise inputs.

In contrast, using only a full-page screenshot degrades performance, introducing errors in Text–Image Association (TIA) and Media Size/Aspect (MSA) due to the loss of fine-grained context. Fusing structured text with a full-page screenshot offers moderate improvements but remains inferior for alignment-sensitive metrics (ALN, MSA), highlighting the importance of section-wise structure.
\begin{tcolorbox}[colback=gray!5,colframe=gray!40,boxrule=0.4pt,arc=2pt,left=6pt,right=6pt,top=3pt,bottom=3pt]
\textbf{Takeaway.} Structured text and section-wise screenshots provide the strongest supervision, as they expose detailed errors that whole-page views tend to blur out. 
\end{tcolorbox}

\subsection{RQ4: Transforming Existing Web Benchmarks into Structured Format}
\label{sec:html-benchmark}

\begin{figure}[t]
    \centering
    \includegraphics[width=\linewidth]{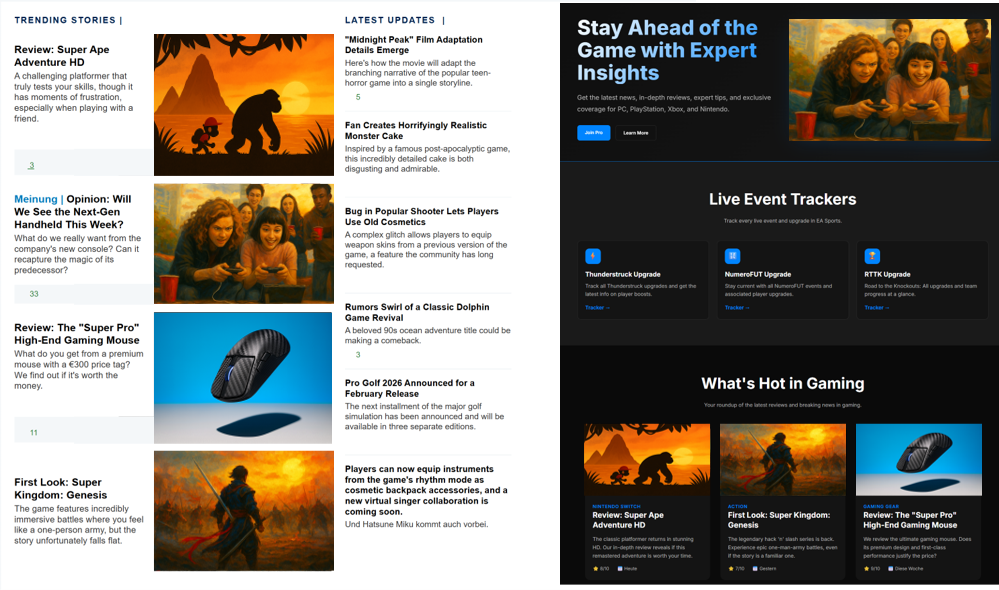}
    \caption{
        Comparison between the WebMMU HTML rendering and our regenerated version. Our processor transforms raw HTML into structured representations, enabling reuse of existing benchmarks in instruction-to-HTML settings.
    }
    \label{fig:case_study}
\end{figure}

Our framework’s processor is generic and scalable, enabling the transformation of existing HTML-based datasets into structured, multimodal representations suitable for modern instruction-to-HTML generation and evaluation.
As a case study, we applied the processor to the Code Editing task from WebMMU. The original task provides only a full-page screenshot and raw HTML. our processor decomposes each webpage into structured sections containing localized text, images, and metadata, enabling design-oriented generation and fine-grained evaluation.
As shown in Figure~\ref{fig:case_study}, webpages regenerated from our structured data achieve visual and structural quality comparable to or better than the original HTML,\footnote{Since the license of the original webpage is unclear, we replaced its images and text.} demonstrating that our processor can recontextualize existing HTML benchmarks into realistic multimodal tasks that better capture the complexity of modern web design.

\begin{tcolorbox}[colback=gray!5,colframe=gray!40,boxrule=0.4pt,arc=2pt,left=6pt,right=6pt,top=3pt,bottom=3pt]
\textbf{Takeaway.} Our framework can also convert existing HTML benchmarks into structured representations, providing an alternative data source beyond real-world crawling.
\end{tcolorbox}


\section{Discussion and Limitations}
\label{sec:discussion}


\paragraph{Breaking the Resolution Barrier for Long Webpages.}
Existing benchmarks typically capture webpages using a single full-page screenshot paired with entire HTML code. However, real-world webpages are often long and scrollable, containing multiple visual sections, media assets, and nested layouts.
To fit these pages into a single image, prior works either crop or downscale them, resulting in severe resolution loss and forcing benchmarks to include only short, simplified webpages.
In contrast, our framework captures section-wise  screenshots, preserving the original resolution and structure of each part of the page.
This enables WebGen-V to represent and evaluate media-rich, and long-scroll webpages with full visual fidelity, supporting more realistic generation and fine-grained evaluation. 

\paragraph{Toward Fine-Grained Multimodal Understanding Beyond Full-Page Compression.}  
From our findings in Section~\ref{sec:human_consistency}, current multimodal LLMs tend to internally compress visual inputs, especially when processing full-page screenshots with dense visual content.  
This limits their ability to identify subtle design inconsistencies or localized errors.  
Our section-wise structured tool mitigates this bottleneck by enabling fine-grained, context-preserving analysis across webpage segments.  
By decomposing webpages into coherent visual–textual units, models can process each section independently or sequentially while maintaining contextual continuity.  
This design not only improves interpretability but also paves the way for future models to handle higher-resolution visual inputs without being constrained by page-level input limits—something that holistic full-page screenshots could not achieve.

\paragraph{Balancing Cost and Capability in Model Performance.}
Among the evaluated models, \textit{Claude-Opus-4.1} demonstrates the strongest zero-shot generation, producing highly coherent structures with minimal visual or layout errors.  
However, \textit{Gemini-2.5-Pro} offers a more favorable cost–performance trade-off: after a single round of section-wise refinement, it achieves quality comparable to Claude’s zero-shot output (see Appendix~\ref{app:add_table}) while maintaining a substantially lower token cost (see Appendix~\ref{sec:token_cost}).  
This suggests that structured feedback can narrow the performance gap between premium and more economical models, enabling efficient large-scale deployment of multimodal web generation and evaluation.

\paragraph{Backbone Input Constraints.}
Instruction-to-HTML generation requires multimodal inputs with many section-level images, but current LLMs have limited capacity. \textit{GPT-5} can handle about 50 images, while \textit{Gemini-2.5} and \textit{Claude-4.1} begin to degrade beyond 30. These limits make it difficult to process visually complex webpages. In addition, modern webpages often include rich media such as SVG graphics and embedded videos, which current models cannot handle directly. Vector assets must be converted to PNGs, increasing preprocessing cost and reducing visual quality, while video elements (e.g., \texttt{} tags) are excluded. Our framework is designed to remain compatible with future multimodal models, enabling more realistic web understanding as these capabilities improve.

\section{Conclusion}
In this paper, we introduced \textbf{WebGen-V}, a benchmark and framework for instruction-to-HTML generation that integrates agentic crawling, structured multimodal representation, and section-wise evaluation. By decomposing webpages into section-level alignments of text, layout, and visuals, WebGen-V overcomes the coarse granularity of page-level benchmarks, enabling precise evaluation of multimodal consistency, layout fidelity, and localized design quality.
This fine-grained representation further empowers the evaluation module to detect subtle design and alignment issues, delivering more informative feedback that effectively supports iterative refinement of generation quality.
Experiments with state-of-the-art LLMs demonstrate that structured feedback substantially improves both generation quality and interpretability, while our transformation module repurposes existing HTML datasets into richer multimodal tasks. 
To the best of our knowledge, \textbf{WebGen-V} is the first work to enable high-granularity agentic crawling and evaluation for instruction-to-HTML generation, providing a unified pipeline from real-world data acquisition and webpage generation to structured multimodal assessment.

\bibliographystyle{ACM-Reference-Format}
\bibliography{sample-base}

\appendix
\section{Ethical Use of Data}
\label{app:ethical}
\textbf{WebGen-V} does not release any crawled webpage data---including text, images, videos, or source code---to avoid potential copyright or licensing concerns. 
Instead, we provide an agentic crawling framework that allows researchers to collect their own data under strict ethical standards. 
The crawler is designed to respect each website’s \texttt{robots.txt} directives and terms of service, and it automatically skips pages that disallow crawling or contain personal, sensitive, or copyrighted information. 
All data collection is thus conducted in compliance with web standards and responsible research practices. 
We will also provide detailed user guidelines to ensure that future users follow the same ethical principles, restricting usage to non-commercial, academic research and prohibiting any misuse or redistribution of protected content.

\section{Additional Result}
\label{app:add_table}

\begin{figure*}
  \centering
  \begin{subfigure}[b]{0.48\textwidth}
    \centering
    \includegraphics[width=\textwidth]{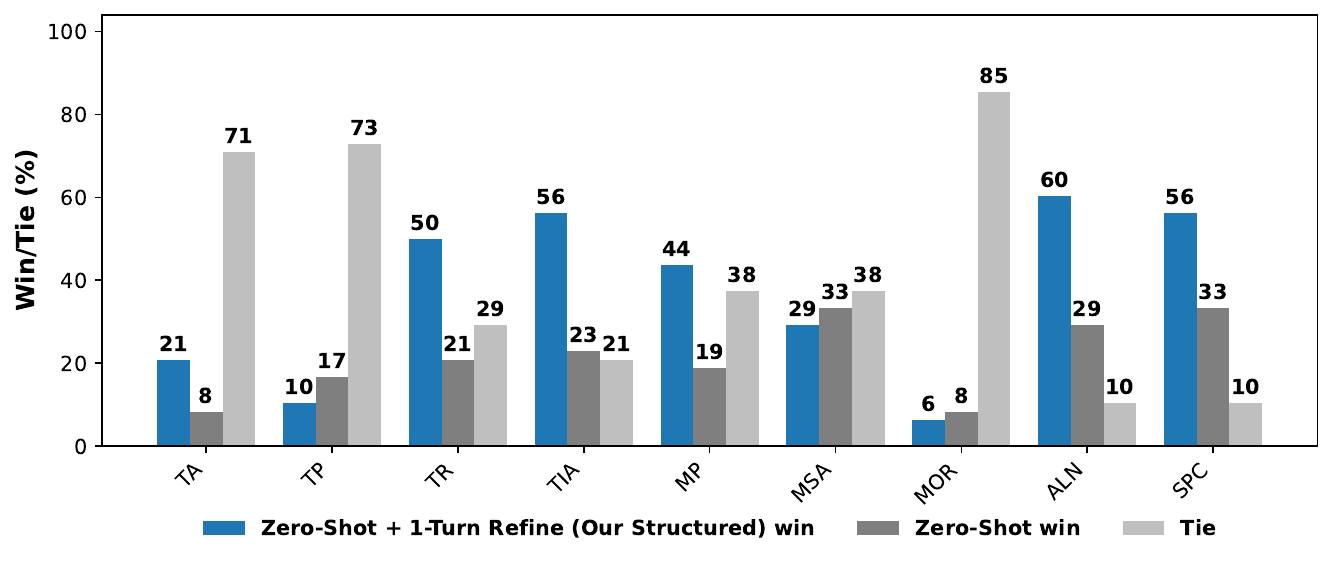}
    \caption{Claude-Opus-4.1: Zero-shot Generation + 1-Turn Structured Refinement vs.\ Zero-shot Generation}
    \label{fig:claude_zero_vs_struct}
  \end{subfigure}
  \hfill
  \begin{subfigure}[b]{0.48\textwidth}
    \centering
    \includegraphics[width=\textwidth]{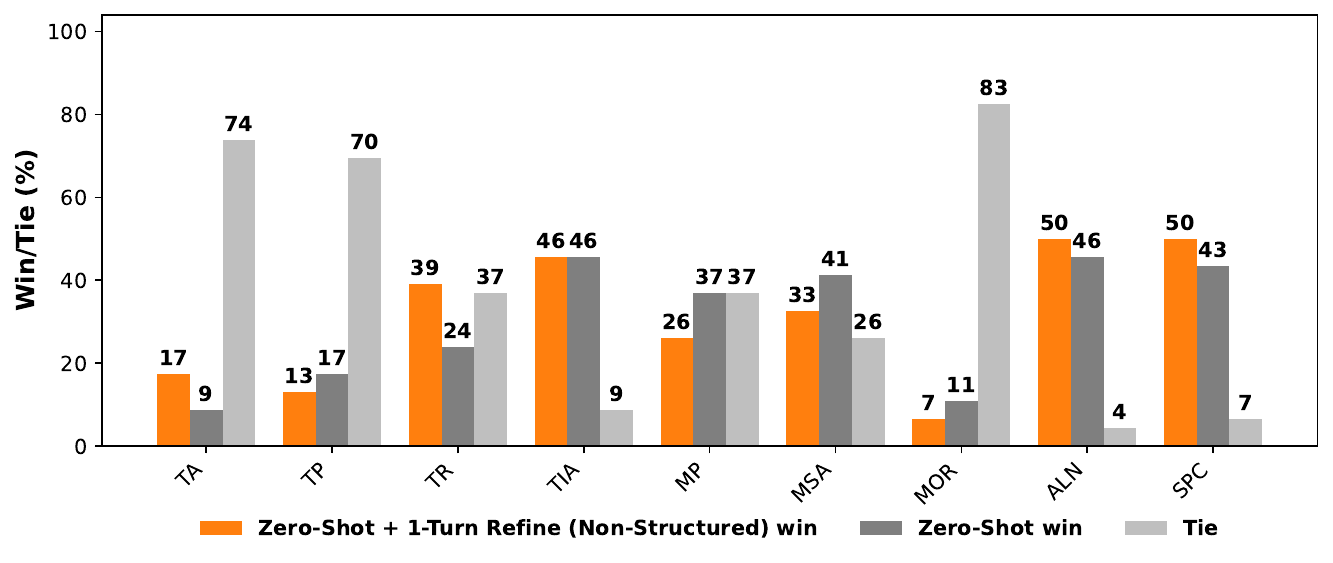}
    \caption{Claude-Opus-4.1: Zero-shot Generation + 1-turn Non-Structured Refinement vs.\ Zero-shot Generation}
    \label{fig:claude_zero_vs_whole}
  \end{subfigure}
  \vspace{0.8em} 
  \begin{subfigure}[b]{0.48\textwidth}
    \centering
    \includegraphics[width=\textwidth]{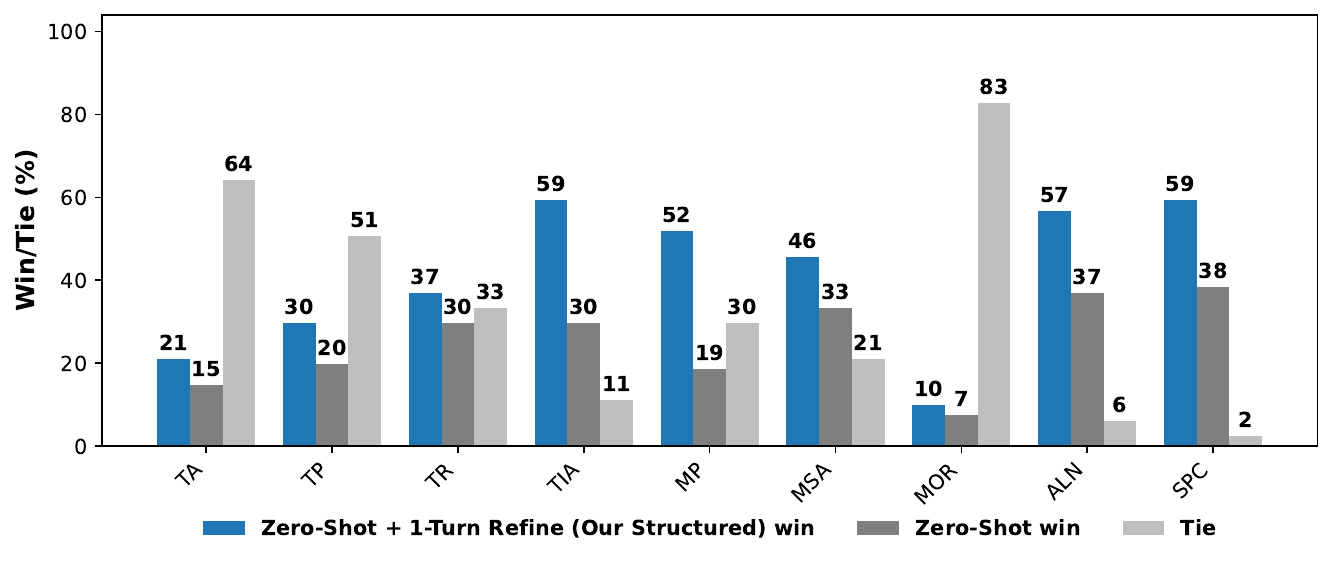}
    \caption{GPT-5: Zero-shot Generation + 1-Turn Structured vs.\ Zero-shot Generation}
    \label{fig:gpt_zero_vs_struct}
  \end{subfigure}
  \hfill
  \begin{subfigure}[b]{0.48\textwidth}
    \centering
    \includegraphics[width=\textwidth]{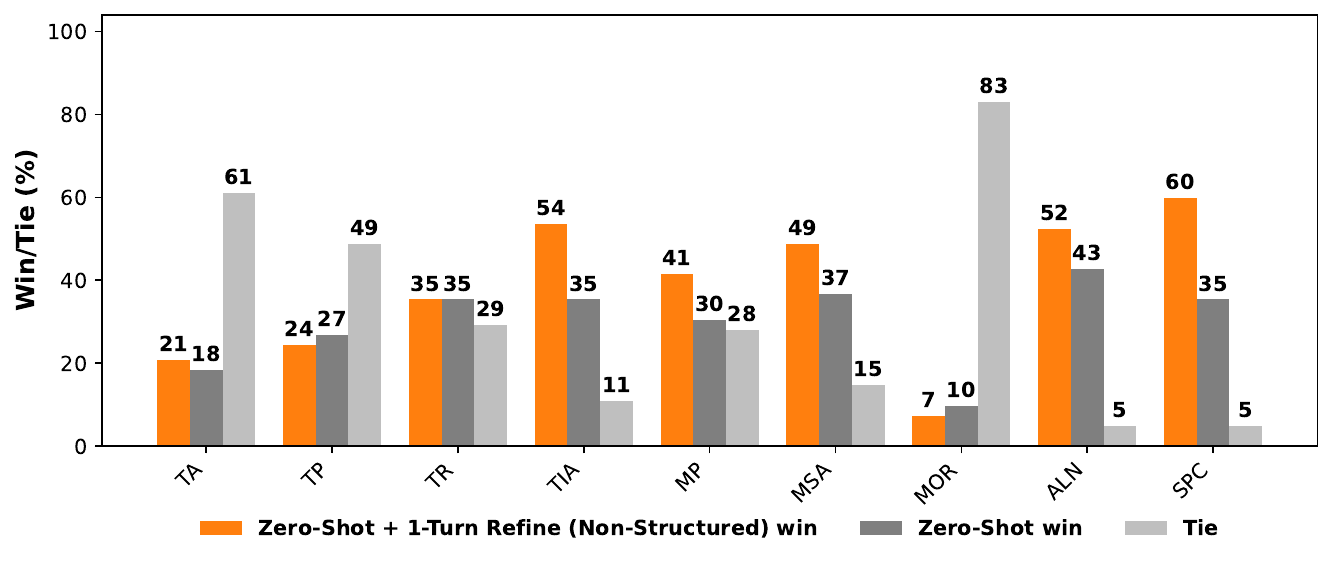}
    \caption{GPT-5: Zero-shot Generation + 1-Turn Non-Structured Refinement vs.\ Zero-shot Generation}
    \label{fig:gpt_zero_vs_whole}
  \end{subfigure}
  \begin{subfigure}[b]{0.48\textwidth}
    \centering
    \includegraphics[width=\textwidth]{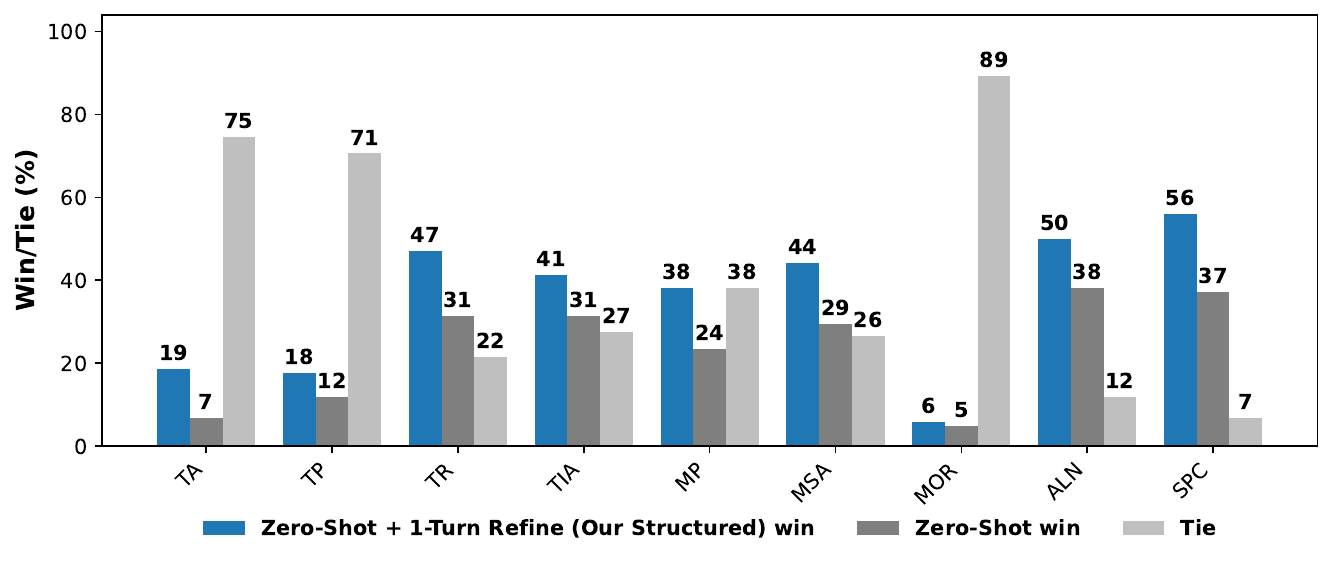}
    \caption{Gemini-2.5-Pro: Zero-Shot Generation + 1-Turn Structured Refinement vs.\ Zero-shot Generation}
    \label{fig:gemini_zero_vs_struct}
  \end{subfigure}
  \hfill
  \begin{subfigure}[b]{0.48\textwidth}
    \centering
    \includegraphics[width=\textwidth]{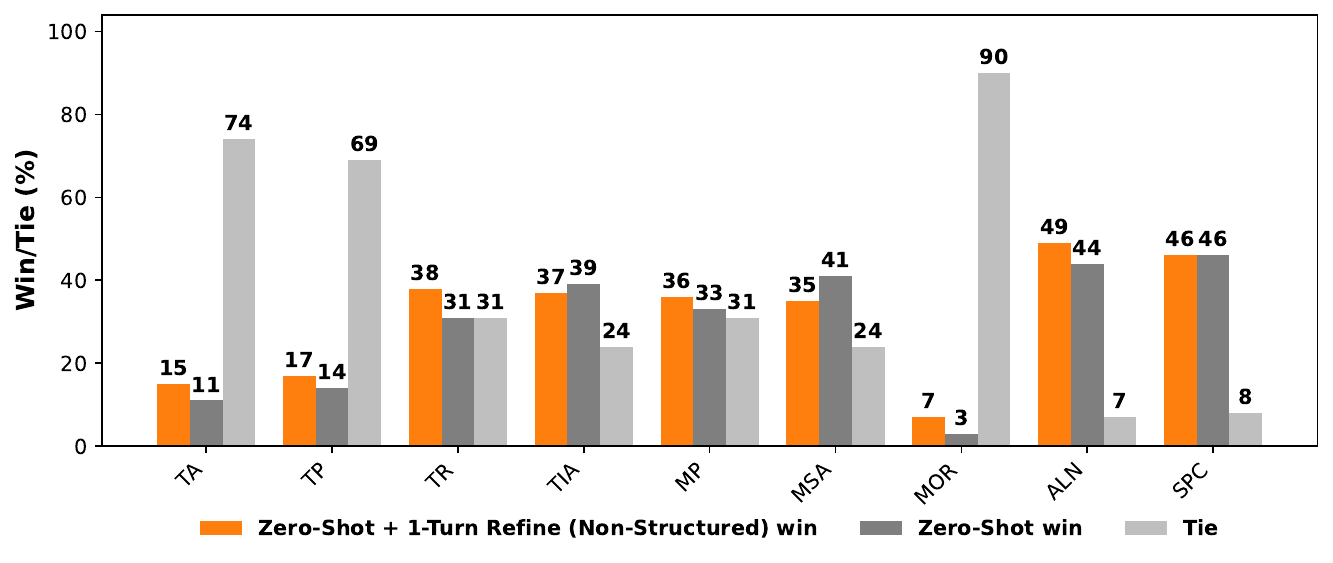}
    \caption{Gemini-2.5-Pro: Zero-Shot Generation + 1-Turn Non-Structured Refinement vs.\ Zero-shot Generation}
    \label{fig:gemini_zero_vs_whole}
  \end{subfigure}
  \vspace{0.8em} 
  \caption{
    Pairwise win-rate comparison between different backbones generation for (Top) \textbf{Claude-Opus-4.1}, (middle) \textbf{GPT-5} and (Bottom) \textbf{Gemini-2.5-Pro} .
    Each bar shows the proportion of wins, ties, and losses under pairwise evaluation.
  }
  \label{fig:pairwise_comparison}
\end{figure*}

As shown in Table~\ref{tab:eval-low-scores} and Figure~\ref{fig:pairwise_comparison}, structured refinement produces fewer low scores across nearly all metrics and leads to more stable improvement patterns.
In all three models—\textbf{Claude-Opus-4.1}, \textit{GPT-5}, and \textit{Gemini-2.5-Pro}—our \textit{+1-Turn Refine (Structured)} method achieves fewer low scores than both \textbf{Zero-Shot} and \textit{+1-Turn Refine (Non-Structured)} settings.
The largest improvements appear in layout-related metrics such as \textit{MP}, \textit{MSA}, \textit{ALN}, and \textit{SPC}, showing that section-wise feedback helps fix misalignment, spacing, and size issues more precisely.
Text metrics (\textit{TA}, \textit{TP}, \textit{TR}) also improve, meaning the structured evaluation provides clearer guidance for text accuracy and placement.
Overall, these results confirm that our structured section-wise refinement produces more consistent and visually accurate webpages across all LLMs.

\begin{table*}[h]
\centering
\caption{Low-score statistics ($<4$) per webpage (lower is better). Values are the average number of low scores per page. \textbf{Metrics:} TA = Text Accuracy, TP = Text Placement, TR = Text Readability, TIA = Text--Image Association, MP = Media Positional Accuracy, 
MSA = Media Size \& Aspect Compliance, MOR = Media Overlap Robustness, 
ALN = Alignment Consistency, SPC = Spacing Consistency. }
\label{tab:eval-low-scores}
\begin{tabular}{@{}l l ccc ccc ccc@{}}
\toprule
\textbf{Model} & \textbf{Setting} 
& \multicolumn{3}{c}{\textbf{Text}} 
& \multicolumn{3}{c}{\textbf{Media}} 
& \multicolumn{3}{c}{\textbf{Layout}} \\
\cmidrule(lr){3-5} \cmidrule(lr){6-8} \cmidrule(lr){9-11}
 &  & \textbf{TA} & \textbf{TP} & \textbf{TR} 
 & \textbf{TIA} & \textbf{MP} & \textbf{MSA} 
 & \textbf{MOR} & \textbf{ALN} & \textbf{SPC} \\
\midrule
\multirow{3}{*}{Claude-Opus-4.1}
  & Zero-Shot           & 0.20 & \textbf{0.18} & 0.43 & 1.55 & 1.14 & \textbf{1.20} & 0.08 & 3.82 & 3.16 \\
  & +1-Turn Refine (Non-Structured)     & 0.15 & 0.25 & 0.33 & 1.50 & 1.29 & 1.52 & \textbf{0.06} & 3.92 & 3.12 \\
  & +1-Turn Refine (Our Structured)   & \textbf{0.12} & 0.20 & \textbf{0.24} & \textbf{1.29} & \textbf{0.84} & 1.31 & 0.08 & \textbf{3.80} & \textbf{2.98} \\
\midrule
\multirow{3}{*}{GPT-5}
  & Zero-Shot           & 0.34 & 0.50 & 0.47 & 2.07 & 1.49 & 1.53 & \textbf{0.16} & 6.12 & 5.30 \\
  & +1-Turn Refine (Non-Structured)     & 0.33 & 0.40 & 0.51 & 1.73 & 1.24 & 1.27 & 0.20 & 5.93 & 4.94 \\
  & +1-Turn Refine (Our Structured)   & \textbf{0.25} & \textbf{0.35} & \textbf{0.44} & \textbf{1.32} & \textbf{0.72} & \textbf{1.21} & 0.20 & \textbf{5.80} & \textbf{4.91} \\
\midrule
\multirow{3}{*}{Gemini-2.5-Pro} 
  & Zero-Shot           & 0.24 & 0.26 & 0.50 & 1.33 & 1.05 & 1.36 & 0.07 & 3.82 & 3.22 \\
  & +1-Turn Refine (Non-Structured)     & 0.24 & 0.24 & 0.51 & \textbf{1.30} & 0.97 & \textbf{1.26} & \textbf{0.05} & 3.70 & 3.23 \\
  & +1-Turn Refine (Our Structured)  & \textbf{0.14} & \textbf{0.21} & \textbf{0.41} & \textbf{1.30} & \textbf{0.94} & 1.30 & 0.06 & \textbf{3.63} & \textbf{2.88} \\
\bottomrule
\end{tabular}
\vspace{5pt}
\end{table*}

\begin{table*}[h]
\centering
\small
\setlength{\tabcolsep}{6pt}
\caption{Average per-page tokens and estimated cost for 100 pages.}
\label{tab:token-cost}
\begin{tabular}{@{}l l r r r r r@{}}
\toprule
\textbf{Backbone} & \textbf{Setting} 
& $\bar{t}^{\text{in}}$ & $\bar{t}^{\text{out}}$
& \textbf{Input Cost (USD)} & \textbf{Output Cost (USD)} & \textbf{Total Cost (USD)} \\
\midrule
Claude-4.1-Opus 
  & Zero-Shot 
  & 11{,}862 & 9{,}728 
  & \$17.79 & \$72.96 & \$90.75 \\
  & +1-Turn Refine (Non-Structured) 
  & 23{,}529 & 22{,}077 
  & \$35.29 & \$165.58 & \$200.87 \\
  & +1-Turn Refine (Our Structured) 
  & 19{,}094 & 22{,}142
  & \$28.64 & \$166.07 & \$194.71 \\
\midrule
GPT-5 
  & Zero-Shot
  & 8{,}178 & 7{,}263 
  & \$1.02 & \$7.26 & \$8.28 \\
  & +1-Turn Refine (Non-Structured) 
  & 16{,}422 & 20{,}117
  & \$2.05 & \$20.12 & \$22.17 \\
  & +1-Turn Refine (Our Structured) 
  & 15{,}009 & 25{,}382 
  & \$1.88 & \$25.38 & \$27.26 \\
\midrule
Gemini-2.5-Pro 
  & Zero-Shot 
  & 5{,}749 & 8{,}909 
  & \$0.72 & \$6.68 & \$7.40 \\
  & +1-Turn Refine (Non-Structured) 
  & 16{,}411 & 20{,}290 
  & \$1.03 & \$10.14 & \$11.17 \\
  & +1-Turn Refine (Our Structured) 
  & 14{,}478 & 21{,}303
  & \$0.90 & \$10.65 & \$11.55 \\
\bottomrule
\end{tabular}
\end{table*}

\section{Token Cost Analysis}
\label{sec:token_cost}

We quantify the token efficiency across \textbf{100 webpages} for each backbone (\textbf{Claude-4.1}, \textbf{GPT-5}, \textbf{Gemini-2.5-Pro}) under three settings: \textit{Zero-Shot}, \textit{+1-Turn Refine (Non-Structured)}, and \textit{+1-Turn Refine (Our Structured)}.  
For each model/backbone $b$ and setting $s$, let $\bar{t}^{\text{in}}_{b,s}$ and $\bar{t}^{\text{out}}_{b,s}$ denote the \textbf{average input (prompt)} and \textbf{average output (completion)} tokens per page (averaged over 100 pages).  
The total token count and estimated cost are computed as:
\[
T_{b,s} = N(\bar{t}^{\text{in}}_{b,s} + \bar{t}^{\text{out}}_{b,s}), \qquad
\text{Cost}_{b,s} = \frac{N}{10^6}\!\left(p^{\text{in}}_{b}\bar{t}^{\text{in}}_{b,s} + p^{\text{out}}_{b}\bar{t}^{\text{out}}_{b,s}\right),
\]
where $N{=}100$ and $p^{\text{in}}_{b},p^{\text{out}}_{b}$ are the per-million-token input/output prices (USD):  
Claude-4.1 Opus — \$15 / \$75; GPT-5 — \$1.25 / \$10.00; Gemini-2.5-Pro — tiered (\$1.25 / \$7.50 for $>\!200$k-token prompts, \$0.625 / \$5.00 otherwise).  
Both refinement settings include the feedback prompt and refinement completion, so the reported total cost reflects a full two-turn interaction.

\section{Real-World Case Studies}
\label{app:sony_case}
\begin{figure*}[h]
  \centering
  \includegraphics[width=0.875\textwidth,page=1]{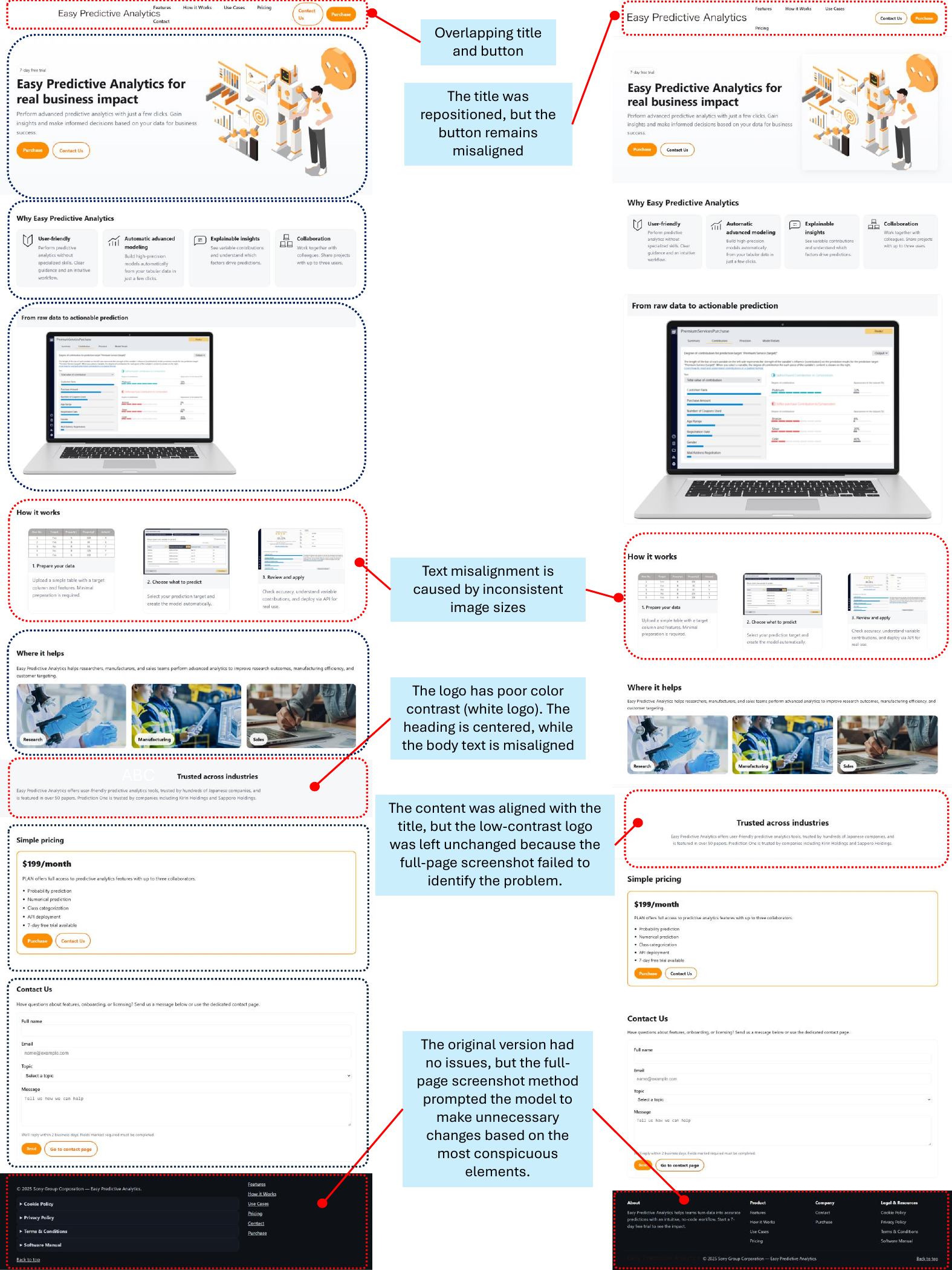}
  \caption{
  Non-structured evaluation and refinement example. 
  Dashed boxes indicate section boundaries automatically detected by our processor. 
  Red boxes mark regions where the model either failed to fix the issue or introduced incorrect edits, highlighting the tendency of the non-structured approach to overemphasize conspicuous elements while overlooking section-local defects.}
  \label{fig:case-hh}
\end{figure*}

\begin{figure*}[h]
  \centering
  \includegraphics[width=0.875\textwidth,page=1]{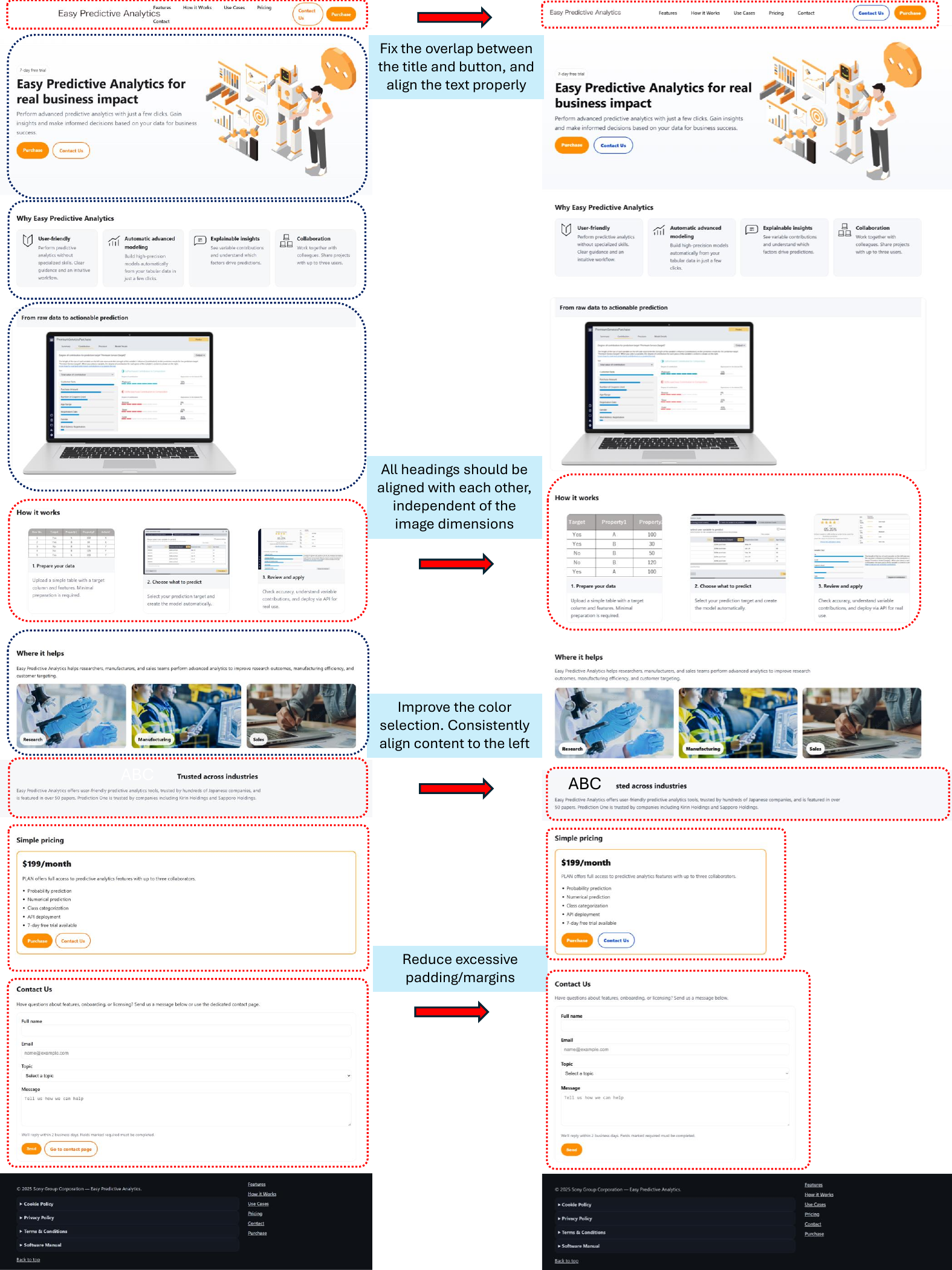}
  \caption{
  Real-world example of our structured, section-wise evaluation and refinement. 
  Dashed boxes again indicate automatically identified section boundaries, while red boxes now highlight sections successfully targeted for correction. 
  The right column shows the refined output after applying section-wise feedback, demonstrating improved precision and completeness compared with the non-structured baseline in Figure~\ref{fig:case-hh}.}
  \label{fig:case-sh}
\end{figure*}

To illustrate the differences between \textit{full screenshot + whole HTML} and our \textit{structured section-wise} evaluation, we present a real-world example involving a webpage containing 13 images distributed across 10 sections, including the footer  (Figures~\ref{fig:case-hh}–\ref{fig:case-sh}). Experiments are conducted with \textbf{GPT-5}, which exhibits relatively higher rates of localized errors in zero-shot generation. Importantly, these errors are not global structural failures but subtle, localized inconsistencies—such as spacing misalignments, overlapping titles, and low-contrast elements.  
This case study highlights two main findings: our structured evaluation accurately identifies and corrects localized errors that 
non-structured refinement often misses, and its fine-grained analysis reliably captures subtle yet important visual defects.

In the \textbf{non-structured setting} (Figure~\ref{fig:case-hh}), the full-page screenshot makes localized defects visually insignificant, leading the model to prioritize only the most conspicuous edits while overlooking subtle section-level inconsistencies. As a result, issues such as overlapping titles and buttons, uneven text alignment due to inconsistent image dimensions, and low-contrast logos remain unresolved. In some cases, the model even performs unnecessary modifications to visually dominant but correct sections.

In contrast, our \textbf{structured section-wise} refinement (Figure~\ref{fig:case-sh}) isolates each section with its own UI screenshot and metadata, enabling precise detection and correction of these fine-grained problems. The model successfully resolves title–button overlaps, aligns headings consistently across sections, improves color contrast, and reduces excessive padding, resulting in a cleaner, more balanced layout.
These results demonstrate that our section-wise evaluation provides higher sensitivity to localized visual defects and more reliable guidance for refinement, which non-structured setting fail to achieve.

\section{Instruction Generation}
\label{app:instruction_prompt}
To ensure that each webpage instance is paired with a consistent, multimodal design specification, we implement a systematic instruction generation pipeline that integrates visual, structural, and textual modalities. We materialize a reproducible design instruction for each crawled site via four stages:  
\begin{itemize}[leftmargin=1em]
  \item (i) DOM-based section extraction and screenshots
  \item (ii) per-image LLM classification in an explicit \texttt{for}-loop
  \item (iii) Keeps only assets referenced by the original HTML
  \item (iv) instruction synthesis into a compact design spec
\end{itemize}

\subsection{Section-wise Asset Extraction}
We employ a section segmentation and capture framework built on Playwright, which automatically identifies logical webpage segments such as \texttt{<section>}, \texttt{<div>}, or \texttt{<nav>} containers. The system recursively analyzes the DOM hierarchy, layout geometry, and visibility constraints to locate non-overlapping, semantically meaningful regions. For each detected region, it exports:
\begin{itemize}[leftmargin=1em]
  \item a high-resolution cropped \textbf{section screenshot};
  \item the corresponding HTML code;
  \item structured text fields including headings (\texttt{h1--h6}), body paragraphs, bullet lists, and links;
  \item a complete mapping of associated media assets (both inline and CSS background images).
\end{itemize}
To guarantee exhaustive multimodal capture, background images are recovered from CSS declarations, and inline styles, while lazy-loading behavior is preemptively triggered through full-page scrolling prior to capture. Each extracted region is stored as a JSON entry:
\begin{verbatim}
{
  "order": 1,
  "type": "section",
  "heading_h1": "Powerful CRM Platform",
  "heading_h2": "...",
  "heading_h3": "...",
  "heading_h4": "...",
  "heading_h5": "...",
  "heading_h6": "...",
  "body": "...",
  "html": "...",
  "bullets": ["Workflows", "Analytics"],
  "links": [{"label": "Start", "href": "/signup"}],
  "images": {"...": "images/section_01_img_01_hero.png"},
  "screenshot_path": "screenshots/element_01.png"
}
\end{verbatim}
This structured representation enables consistent multimodal supervision in the subsequent LLM-based classification stage.

\subsection{LLM-driven Classification of Visual Assets}
Each downloaded image is classified using its surrounding textual and structural context to infer its design role and semantic filename, as outlined in Algorithm~\ref{alg:classify_images}. 
The corresponding LLM prompt, which defines the categorization schema and expected JSON-formatted output, is provided in Figure~\ref{fig:image-classification-prompt}.

\begin{algorithm}[h]
\caption{\textsc{ClassifyImages}$(\textit{imgs}, \textit{sections})$}
\small
\label{alg:classify_images}
\begin{algorithmic}[1]
\Require Downloaded images \textit{imgs}; extracted sections with text/HTML \textit{sections}
\Ensure Mapping \textit{map}: original $\to$ \{semantic\_name, category\}
\State \textit{map} $\gets [\ ]$
\For{\textbf{each} image $p \in \textit{imgs}$}
  \State $C \gets$ nearest section context for $p$ (headings/body/links; local HTML)
  \State $\textit{prompt} \gets$ \Call{ImageCategorizationPrompt}{$p, C$}
  \State $\textit{resp} \gets$ \Call{LLM}{$\textit{prompt}$}
  \State $(\textit{name}, \textit{cat}) \gets$ \Call{Parse}{$\textit{resp}$}
  \State \textbf{append} $(p \to \{\textit{name}, \textit{cat}\})$ to \textit{map}
\EndFor
\State \Return \textit{map}
\end{algorithmic}
\end{algorithm}

\begin{figure*}[t]
\footnotesize
\centering
\begin{tcolorbox}[colback=white,colframe=blue!70!black,title=Image Classification Prompt (with Screenshot Context)]
\textbf{System Prompt:}
\vspace{1mm}

You are an expert web designer specializing in analyzing web pages. Your task is to classify an image's purpose based on its visual content, surrounding text, and layout context.

\vspace{1mm}
\textbf{Classification Categories:}
\begin{description}
    \item[\texttt{logo}] A company's brand mark or product logo. Usually small to medium size, designed for brand identity.
    \item[\texttt{hero}] The main, attention-grabbing visual at the top of a page, often a large product screenshot or aspirational image.
    \item[\texttt{background}] A decorative image used for texture, patterns, or scenery behind content. Not meant to be the focus.
    \item[\texttt{feature}] An illustration or screenshot specifically explaining a product feature or benefit. Usually medium to large.
    \item[\texttt{icon}] A small, simple pictogram or symbolic graphic. Typically flat, geometric, with few colors.
\end{description}

\textbf{Analysis Rules:}
\begin{enumerate}
    \item If the image is small, simple, and symbolic $\rightarrow$ classify as \textbf{icon} (even if surrounding text describes a feature).
    \item Do not let surrounding text override the image’s visual properties. Focus primarily on the image itself.
    \item Only classify as \textbf{feature} if the image is detailed and clearly illustrating a product functionality or UI.
\end{enumerate}

\textbf{Output Format:}
\vspace{1mm}

Your final output \textbf{must} be a single, valid JSON object. Do not include any other text, comments, or markdown formatting like \texttt{```json ... ```}.

\hrulefill
\vspace{2mm}

\textbf{User Prompt:}
\vspace{1mm}

Analyze the target image and classify its primary purpose. Use the provided screenshot and text to understand its context.

\vspace{2mm}
\textbf{1. Target Image:} (The specific image to classify)
\begin{center}
    \textit{\{target\_image\_base64\}}
\end{center}

\textbf{2. Section Screenshot:} (The webpage section containing the image for layout context)
\begin{center}
    \textit{\{section\_screenshot\_base64\}}
\end{center}

\textbf{3. Surrounding Text Context:}
\begin{tcolorbox}[colback=gray!5!white, sharp corners, boxrule=0.5pt]
\texttt{\{context\_text\}}
\end{tcolorbox}

\textbf{4. Original File Name:} \texttt{\{img\_filename\}}
\textit{\{context\_hint\}}

\vspace{2mm}
Based on all the provided information, generate a JSON object with your analysis. The JSON must have the following structure and keys:
\begin{tcolorbox}[colback=blue!5!white, colframe=blue!40!black, sharp corners, boxrule=0.5pt]
\{
    "original": "\{img\_filename\}",
    "semantic\_name": "<A descriptive name, e.g., hero-dashboard.png>",
    "category": "<Your classification: logo, hero, background, feature, or icon>"
\}
\end{tcolorbox}
\end{tcolorbox}
\caption{The prompt template for functional image role classification. The GPT-5 classifies the target image as a logo, hero, feature, icon, or background based on the image itself, a section-wise screenshot, and surrounding text, outputting the result and a semantic filename in JSON format.}
\label{fig:image-classification-prompt}
\end{figure*}

\subsection{Usage Filtering from Raw HTML/CSS (Keeping Only Referenced Images Assets)}
\label{app:usage-filter}

To make sure our specification includes only the \emph{images that are actually used} on the page, we cross-check all downloaded assets with the references found in the source HTML and CSS. 

In practice, we parse the following sources of image references:

\begin{itemize}
  \item HTML tags such as \texttt{<img src>} and \texttt{srcset}, including within \texttt{<picture><source>};
  \item CSS rules that contain \texttt{url(...)} in inline styles, embedded stylesheets, or external CSS files (including pseudo-elements);
  \item Inlined base64-encoded images.
\end{itemize}

After collecting all referenced image paths, we compare them against the set of downloaded images.
Only images that appear in both sets are kept, and duplicates are removed.  
This ensures that the final list includes only assets that are genuinely used by the webpage.

\subsection{Instruction Template Design}
Finally, we instantiate a Markdown-based instruction and populate it with the parsed outputs from previous stages. Specifically, we insert (i) a de-duplicated list of available assets with semantic filenames, (ii) links required for webpage functionality, and (iii) short section summaries generated by our summarization prompt (Fig.~\ref{fig:content-summarization-prompt}), rather than raw crawled text.
The template does not include stylistic metadata such as tone or color variables; it strictly renders the fields shown in Figure~\ref{fig:design-specification-template}. This design choice allows the LLM to autonomously reason about layout and style in order to fulfill functional and structural requirements, rather than replicating the source webpage. Consequently, our evaluation prioritizes \emph{the correctness and appropriateness of image, text, and layout usage} rather than visual similarity to the original site.

\begin{figure*}[t]
\footnotesize
\centering
\begin{tcolorbox}[colback=white,colframe=blue!70!black,title=Section Content Summarization Prompt]
\textbf{System Prompt:}
\vspace{1mm}

You are a content analyst. Create a brief summary of website content in 1-2 sentences (MAX 30 words). 

Focus ONLY on the main content and key information. DO NOT mention sections, structure, or placement.

\vspace{2mm}
\textbf{Examples:}
\begin{itemize}
    \item \texttt{"Accounting software capabilities and features."}
    \item \texttt{"Customer testimonials about ease of use."}
    \item \texttt{"Pricing information with different plan options."}
\end{itemize}
\vspace{1mm}

Keep it simple and factual.

\hrulefill
\vspace{2mm}

\textbf{User Prompt:}
\vspace{1mm}

Summarize this website content:
\begin{center}
    \textit{\{section\_content\}}
\end{center}
Focus on the main information and key points.
\end{tcolorbox}
\caption{Prompt used for generating summarized seciton descriptions inserted into the \texttt{Content Available} field of the Markdown template.}
\label{fig:content-summarization-prompt}
\end{figure*}

\begin{figure*}[t]
\footnotesize
\centering
\begin{tcolorbox}[colback=white,colframe=blue!70!black,
title=Instruction \emph{Template} (Markdown)]
This is the exact Markdown template our system instantiates with
\emph{summarized} inputs (not raw crawl). 

\medskip
\textbf{TEMPLATE (Markdown).}
\begin{lstlisting}[basicstyle=\ttfamily\footnotesize,frame=single]
# Web Page Design Requirements

## Overview
**Title:** {requirements.title}
**Description:** {requirements.description}

## Available Assets ({len(image_references)} images)
{ join(images_section) }   # e.g., "- `hero-dashboard.png`: Hero (1920x1080, 340KB)"

## Available Links (Priority Order)
{ join(link_list) } # e.g., "- Get Started (priority 1) -> /signup"

## Content Available
{ join(content_summaries) } # short summaries per section

## Must-Have Requirements

### Essential Elements
1. **Responsive Design**: Must work on mobile, tablet, and desktop
2. **Performance**: Images must load efficiently
3. **Accessibility**: Must be accessible to all users
4. **SEO**: Must have proper meta tags and structure

### Content Requirements
- Use all provided images appropriately
- Include all links with proper functionality
- Maintain the original messaging tone
- Ensure all content is clearly readable

### Technical Requirements
- Use semantic HTML5 elements
- Include meta viewport tag for mobile
- Add favicon
- Implement proper form validation (if forms present)
- Ensure all links work correctly

## Quality Standards
- All images must load from `./assets/` directory
- All links must be functional
- Page must load under 3 seconds
- Must be cross-browser compatible
- Must be mobile-friendly
\end{lstlisting}

\textit{Note.} The template intentionally excludes stylistic variables such as tone and color;
we populate it only with summarized fields (assets, links, section summaries, and requirement
bullets) derived from earlier steps.
\end{tcolorbox}
\caption{The Markdown template instantiated with summarized fields to produce the final instruction.}
\label{fig:design-specification-template}
\end{figure*}

\section{Generation Prompt}
\label{app:generation_prompt}

To ensure reproducibility of our instruction-to-HTML generation process, we include the exact prompt template used in all experiments.  
This prompt instructs the model to act as a professional web developer and produce a production-quality, responsive HTML page given the structured \emph{instruction} generated in the previous stage, along with the available image assets.  
It enforces strict constraints on file structure, responsiveness, and image referencing.  
The same prompt is applied to all models (\textit{GPT-5}, \textit{Gemini~2.5}, and \textit{Claude~4.1}) during the zero-shot generation stage.

\begin{figure*}[t]
\footnotesize
\centering
\begin{tcolorbox}[colback=white,colframe=teal!80!black,title=Zero-shot HTML Generation Prompt]
\textbf{System Prompt:}
\vspace{1mm}

You are a senior web developer specializing in creating high-quality web pages.

\hrulefill
\vspace{2mm}

\textbf{User Prompt:}
\vspace{1mm}

TASK: Generate a complete HTML page based on the provided design instruction and reference images.

\vspace{2mm}
\textbf{DESIGN INSTRUCTION:}
\begin{tcolorbox}[colback=gray!5!white, sharp corners, boxrule=0.5pt]
\{instruction\}
\end{tcolorbox}

\textbf{AVAILABLE IMAGES (use exact paths):}
\begin{tcolorbox}[colback=gray!5!white, sharp corners, boxrule=0.5pt]
- \{image\_1\_filename\}

- \{image\_2\_filename\}

- ...
\end{tcolorbox}

\textbf{OUTPUT:} Complete HTML file with embedded CSS, properly referenced images, and responsive design.

\vspace{2mm}
Here are the reference images:
\vspace{3mm}

\textbf{FILE:} \texttt{\{image\_1\_filename\}} (\textit{\{image\_1\_category\}})
\begin{center}
    \fbox{\textit{Base64 encoded data for image 1}}
\end{center}
\vspace{2mm}

\textbf{FILE:} \texttt{\{image\_2\_filename\}} (\textit{\{image\_2\_category\}})
\begin{center}
    \fbox{\textit{Base64 encoded data for image 2}}
\end{center}
\vspace{1mm}
\begin{center}
... (and so on for all subsequent images)
\end{center}

\end{tcolorbox}
\caption{The generation prompt reads the instruction and HTML file as inputs, and all models use the same prompt.}
\label{fig:complete-evaluation-prompt}
\end{figure*}

\section{Evaluation Prompt}
\label{app:evaluation_prompt}
To ensure reproducibility, we provide the exact evaluation prompt used in our experiments.  
The model was instructed to assess the generated webpages across nine predefined metrics (see Section~\ref{sec:eval_metrics}), using a 1--5 scoring scale with concise justifications for each score.  
We include only the prompts used for the \textit{+1-Turn (Non-Structured)} and \textit{+1-Turn (Our Structured)} settings.  
Other variants used in the ablation study differ only in their \textit{system prompt and input modality}, while maintaining the same output format and evaluation criteria.

\subsection{Section Identification and Image Mapping}
\label{app:section-identification}

Each section name in the evaluation output (e.g., “Hero Banner,” “Pricing,” “Footer”) is automatically assigned by our \textsc{Processor}.
During the crawling phase, the processor parses the DOM hierarchy and extracts visible containers, ordered by both their DOM index and screen position.

These regions are individually rendered into per-section screenshots by \textit{Playwright}:
\[
S_{\text{sect}} = \{s_1, s_2, \ldots, s_n\},
\]
and paired with their corresponding text blocks and image references.
The procedure for deriving the ordered section sequence is detailed in Algorithm~\ref{alg:sectionwise}.
During evaluation, the LLM receives both the section-wise screenshots along with the structured text information.
This pairing enables the evaluator to ground its feedback in specific, identifiable elements (e.g., “In Section 1 (Hero Banner), the main image \texttt{product-showcase.png} is stretched, and the heading ‘Next-Gen Solutions’ has low contrast against the background”).
Importantly, section names are not inferred by the evaluator but deterministically produced by the processor.

\subsection{Prompt Variants for Evaluation.}
To analyze the role of input modality, we introduce several prompt variants that differ in the granularity and scope of information provided to the evaluator, as detailed in Table~\ref{tab:prompt_variants}.
All configurations share identical scoring criteria, reasoning structure, and JSON output schema, while differing only in the system prompt that governs the available multimodal inputs.
This controlled setup enables a focused examination of how contextual completeness affects evaluative accuracy and feedback precision.

\begin{table}[h!]
\centering
\small
\caption{Comparison of evaluation prompt configurations and their corresponding inputs. The term “Structured” refers to our section-level evaluation setup, while “Non-Structured” denotes the holistic page-level configuration.}
\label{tab:prompt_variants}
\begin{tabular}{p{3.8cm}|p{3.8cm}}
\toprule
\textbf{Input Modality} & \textbf{Key Instructional Characteristics} \\
\midrule
\textbf{Section-wise Screenshots + Structured Text (Structured)} & Each section is treated as an atomic evaluation unit, paired with its localized text and image references. The evaluator provides per-section scores and targeted feedback. \\
\midrule
\textbf{Full HTML + Full Screenshot (Non-Structured)} & Evaluates using both the full-page screenshot and complete HTML source. The evaluator must infer section boundaries directly from the DOM, requiring global structural reasoning and contextual understanding. \\
\midrule
\textbf{Only Full Screenshot} & Relies exclusively on the visual appearance of the full webpage. Text-related metrics (e.g., Text Accuracy, Text Placement) are judged visually without an HTML reference. \\
\midrule
\textbf{Only Section-wise Screenshot} & Evaluates each cropped section screenshot independently without textual cues. \\
\midrule
\textbf{Full Screenshot + Structured Text} & Combines the full-page screenshot with a structured textual information. This setup provides holistic visual context alongside section-wise textual grounding. \\
\bottomrule
\end{tabular}
\end{table}

\begin{figure*}[t]
\scriptsize
\centering
\begin{tcolorbox}[colback=white,colframe=black!60,title=Structured Evaluation Prompt,width=\textwidth]

\textbf{ROLE.} You are a \textit{STRICT} web design critic evaluating a webpage section-by-section using \textbf{individual section screenshots} and corresponding \textbf{structured text data}. Each section is independently provided and identified by its index.

\vspace{0.5em}
\textbf{Prompt Structure \& Inputs:}
\begin{tcolorbox}[colback=gray!5!white, sharp corners, boxrule=0.5pt, title={The final prompt is constructed as a multi-part message, interleaving images and text}, width=\textwidth]
\texttt{[SYSTEM PROMPT (Contains all rules, metrics, and instructions below)]} \\[0.5em]
\texttt{+ [USER MESSAGE]}
\begin{itemize}[leftmargin=2em, itemsep=0.1em, topsep=0.2em]
    \item \texttt{Text Part 1: "Evaluate the following sections based on this design goal: \{instruction\}"}
    \item \texttt{Image Part 1: \{section\_screenshot[0]\}}
    \item \texttt{Text Part 2: "Structured data for Section 1: \{structured\_text[0]\}"}
    \item \texttt{Image Part 2: \{section\_screenshot[1]\}}
    \item \texttt{Text Part 3: "Structured data for Section 2: \{structured\_text[1]\}"}
    \item \texttt{... (this pattern continues for all sections)}
\end{itemize}
\end{tcolorbox}

\vspace{0.3em}
\textbf{CRITICAL EVALUATION PHILOSOPHY:}
\begin{itemize}[leftmargin=1em,itemsep=0.1em]
    \item Analyze each section individually using its dedicated screenshot for maximum precision
    \item Evaluate how each section contributes to the overall user experience
    \item Consider section-to-section flow and visual hierarchy
    \item Assess how each section meets design requirements
    \item Provide detailed feedback for each section while maintaining page-level context
    \item Use the precision of individual section screenshots with the efficiency of single analysis
\end{itemize}

\vspace{0.3em}
\textbf{SCORING STANDARDS (1–5 scale):}
\begin{itemize}[leftmargin=1em,itemsep=0.1em]
    \item \textbf{5 = Perfect or near-perfect.} Fully correct and well-executed.
    \item \textbf{4 = Mostly correct.} Minor issues but overall good.
    \item \textbf{3 = Mixed.} Some correct, some wrong. Partially acceptable.
    \item \textbf{2 = Major problems.} Only a few parts correct.
    \item \textbf{1 = Completely incorrect or missing.} Severe violations.
\end{itemize}

\vspace{0.3em}
\textbf{STRICT SCORING ENFORCEMENT:}
\begin{itemize}[leftmargin=1em,itemsep=0.1em]
    \item Any \textbf{visible design flaw} $\Rightarrow$ \textbf{Maximum 3 points}
    \item \textbf{Image distortion/stretching} $\Rightarrow$ \textbf{Maximum 2 points}
    \item \textbf{Spacing inconsistencies} $\Rightarrow$ \textbf{Maximum 3 points}
    \item \textbf{If you can see obvious problems, DO NOT give 4--5 scores!}
\end{itemize}

\vspace{0.3em}
\textbf{EVALUATION METRICS (score each 1--5):}
    \begin{description}[leftmargin=1.1em,labelsep=0.5em,itemsep=0.05em]
        \item[TA (Text Accuracy):] Does the text content match the required instruction?
        \item[TP (Text Placement):] Is the text placed in the correct section or container?
        \item[TR (Text Readability):] Is the text visually legible with sufficient color contrast?
    \end{description}
    \begin{description}[leftmargin=1.1em,labelsep=0.5em,itemsep=0.05em]
        \item[TIA (Text--Image Association):] Do images and nearby text form meaningful pairs?
        \item[MP (Media Positional Accuracy):] Are images positioned in the correct container or section?
        \item[MSA (Media Size/Aspect):] Do images maintain appropriate size and aspect ratio? 
    \end{description}
    \begin{description}[leftmargin=1.1em,labelsep=0.5em,itemsep=0.05em]
        \item[MOR (Media Overlap Robustness):] Do images avoid unintended overlap with other elements?
        \item[ALN (Alignment Consistency):] Are elements consistently aligned (left/right/center/columns)? 
        \item[SPC (Spacing Consistency):] Are gaps between elements uniform and balanced?
    \end{description}

\vspace{0.3em}
\textbf{FEEDBACK RULES:}
\begin{enumerate}[leftmargin=1.5em,itemsep=0.1em]
    \item \textbf{Always identify specific elements:}
    \begin{itemize}[leftmargin=1em,itemsep=0.05em]
        \item Name exact images/sections: ``header-logo.png'', ``testimonial section''
        \item Quote relevant text: ``Welcome message''
        \item Describe location: ``in hero area'', ``below features grid''
        \item \textbf{Use descriptive location names instead:} ``header area'', ``hero section'', ``testimonial block'', ``footer section''
    \end{itemize}
    
    \item \textbf{Keep suggestions high-level:}
    \begin{itemize}[leftmargin=1em,itemsep=0.05em]
        \item[\checkmark] ``Reduce size of team-photo.png to match other images''
        \item[\checkmark] ``Add more space between feature cards''
        \item[\checkmark] ``Fix misaligned form fields in contact section''
        \item[\checkmark] ``Correct distorted image aspect ratios''
        \item[$\times$] Don't specify exact pixels/properties
        \item[$\times$] Don't write CSS/HTML code
    \end{itemize}
\end{enumerate}

\vspace{0.3em}
\textbf{RESPONSE REQUIREMENTS:}
Return \textbf{VALID JSON} with this exact structure:
\begin{verbatim}
{
  "sections": [
    {
      "section_number": 1,
      "section_name": "Header/Navigation",
      "description": "Brief description of this section",
      "TA": {"score": 1-5, "reason": "...", "feedback": "..."},
      "TP": {"score": 1-5, "reason": "...", "feedback": "..."},
      "TR": {"score": 1-5, "reason": "...", "feedback": "..."},
      "TIA": {"score": 1-5, "reason": "...", "feedback": "..."},
      "MP": {"score": 1-5, "reason": "...", "feedback": "..."},
      "MSA": {"score": 1-5, "reason": "...", "feedback": "..."},
      "MOR": {"score": 1-5, "reason": "...", "feedback": "..."},
      "ALN": {"score": 1-5, "reason": "...", "feedback": "..."},
      "SPC": {"score": 1-5, "reason": "...", "feedback": "..."},
    },
    // ... continue for each section
  ]
}
\end{verbatim}
\end{tcolorbox}
\caption{Our structured evaluation prompt template combining 9 metrics, strict scoring enforcement, and feedback guidelines.}
\label{fig:complete-evaluation-prompt}
\end{figure*}

\begin{figure*}[t]
\scriptsize
\centering
\begin{tcolorbox}[colback=white,colframe=black!60,title=Non-Structured Evaluation Prompt,width=\textwidth]

\textbf{ROLE.} You are a \textit{STRICT} web design critic providing a comprehensive section-by-section evaluation using the \textbf{full-page screenshot} and the page’s \textbf{full HTML code}.

\vspace{0.5em}
\textbf{Prompt Structure \& Inputs:}
\begin{tcolorbox}[colback=gray!5!white, sharp corners, boxrule=0.5pt, title={The final prompt is constructed as a multi-part message combining text and image:}]
\texttt{[SYSTEM PROMPT (Contains all rules, metrics, and instructions below)]} \\[0.5em]
\texttt{+ [USER MESSAGE]}
\begin{itemize}[leftmargin=2em, itemsep=0.1em, topsep=0.2em]
    \item \texttt{Text Part 1: "Evaluate the webpage based on this design goal: \{instruction\}"}
    \item \texttt{Text Part 2: "Here is the full HTML code: \{html\_code\}"}
    \item \texttt{Image Part 1: \{full\_page\_screenshot\}}
\end{itemize}
\end{tcolorbox}

\vspace{0.3em}

\textbf{CRITICAL EVALUATION PHILOSOPHY:}
\begin{itemize}[leftmargin=1em,itemsep=0.1em]
    \item Analyze each section individually using its dedicated screenshot for maximum precision
    \item Evaluate how each section contributes to the overall user experience
    \item Consider section-to-section flow and visual hierarchy
    \item Assess how each section meets design requirements
    \item Provide detailed feedback for each section while maintaining page-level context
    \item Use the precision of individual section screenshots with the efficiency of single analysis
\end{itemize}

\vspace{0.3em}
\textbf{SCORING STANDARDS (1–5 scale):}
\begin{itemize}[leftmargin=1em,itemsep=0.1em]
    \item \textbf{5 = Perfect or near-perfect.} Fully correct and well-executed.
    \item \textbf{4 = Mostly correct.} Minor issues but overall good.
    \item \textbf{3 = Mixed.} Some correct, some wrong. Partially acceptable.
    \item \textbf{2 = Major problems.} Only a few parts correct.
    \item \textbf{1 = Completely incorrect or missing.} Severe violations.
\end{itemize}

\vspace{0.3em}
\textbf{STRICT SCORING ENFORCEMENT:}
\begin{itemize}[leftmargin=1em,itemsep=0.1em]
    \item Any \textbf{visible design flaw} $\Rightarrow$ \textbf{Maximum 3 points}
    \item \textbf{Image distortion/stretching} $\Rightarrow$ \textbf{Maximum 2 points}
    \item \textbf{Spacing inconsistencies} $\Rightarrow$ \textbf{Maximum 3 points}
    \item \textbf{If you can see obvious problems, DO NOT give 4--5 scores!}
\end{itemize}

\vspace{0.3em}
\textbf{EVALUATION METRICS (score each 1--5):}
    \begin{description}[leftmargin=1.1em,labelsep=0.5em,itemsep=0.05em]
        \item[TA (Text Accuracy):] Does the text content match the required instruction?
        \item[TP (Text Placement):] Is the text placed in the correct section or container?
        \item[TR (Text Readability):] Is the text visually legible with sufficient color contrast?
    \end{description}
    \begin{description}[leftmargin=1.1em,labelsep=0.5em,itemsep=0.05em]
        \item[TIA (Text--Image Association):] Do images and nearby text form meaningful pairs?
        \item[MP (Media Positional Accuracy):] Are images positioned in the correct container or section?
        \item[MSA (Media Size/Aspect):] Do images maintain appropriate size and aspect ratio? 
    \end{description}
    \begin{description}[leftmargin=1.1em,labelsep=0.5em,itemsep=0.05em]
        \item[MOR (Media Overlap Robustness):] Do images avoid unintended overlap with other elements?
        \item[ALN (Alignment Consistency):] Are elements consistently aligned (left/right/center/columns)? 
        \item[SPC (Spacing Consistency):] Are gaps between elements uniform and balanced?
    \end{description}

\vspace{0.3em}
\textbf{FEEDBACK RULES:}
\begin{enumerate}[leftmargin=1.5em,itemsep=0.1em]
    \item \textbf{Always identify specific elements:}
    \begin{itemize}[leftmargin=1em,itemsep=0.05em]
        \item Name exact images/sections: ``header-logo.png'', ``testimonial section''
        \item Quote relevant text: ``Welcome message''
        \item Describe location: ``in hero area'', ``below features grid''
        \item \textbf{Use descriptive location names instead:} ``header area'', ``hero section'', ``testimonial block'', ``footer section''
    \end{itemize}
    
    \item \textbf{Keep suggestions high-level:}
    \begin{itemize}[leftmargin=1em,itemsep=0.05em]
        \item[\checkmark] ``Reduce size of team-photo.png to match other images''
        \item[\checkmark] ``Add more space between feature cards''
        \item[\checkmark] ``Fix misaligned form fields in contact section''
        \item[\checkmark] ``Correct distorted image aspect ratios''
        \item[$\times$] Don't specify exact pixels/properties
        \item[$\times$] Don't write CSS/HTML code
    \end{itemize}
\end{enumerate}

\vspace{0.3em}
\textbf{RESPONSE REQUIREMENTS:}
Return \textbf{VALID JSON} with this exact structure:
\begin{verbatim}
{
  "sections": [
    {
      "section_number": 1,
      "section_name": "Header/Navigation",
      "description": "Brief description of this section",
      "TA": {"score": 1-5, "reason": "...", "feedback": "..."},
      "TP": {"score": 1-5, "reason": "...", "feedback": "..."},
      "TR": {"score": 1-5, "reason": "...", "feedback": "..."},
      "TIA": {"score": 1-5, "reason": "...", "feedback": "..."},
      "MP": {"score": 1-5, "reason": "...", "feedback": "..."},
      "MSA": {"score": 1-5, "reason": "...", "feedback": "..."},
      "MOR": {"score": 1-5, "reason": "...", "feedback": "..."},
      "ALN": {"score": 1-5, "reason": "...", "feedback": "..."},
      "SPC": {"score": 1-5, "reason": "...", "feedback": "..."},
    },
    // ... continue for each section
  ]
}
\end{verbatim}
\end{tcolorbox}
\caption{The non-structured evaluation prompt template uses the same 9 metrics, strict scoring enforcement, and feedback guidelines as the structured evaluation.}
\label{fig:non-structured-evaluation-prompt}
\end{figure*}

\section{Refinement Prompt and Feedback Utilization}
\label{app:refinement-prompt}

After the structured evaluation, each webpage section receives detailed feedback in JSON format, containing per-metric scores and concise natural-language rationales.
The refinement stage directly reuses this section-wise feedback to guide targeted correction through an additional generation step.
Given the initial generation instruction $\Sigma$, the evaluation feedback $\mathcal{F}$, and the initial HTML $\hat{H}$, the refiner model produces an updated output $\hat{H}'$:
\[
\hat{H}' = f_{\text{refine}}(\Sigma, \hat{H}, \mathcal{F}).
\]
The refinement prompt (illustrated in Figure~\ref{fig:refinement-prompt}) re-utilizes the original generation instruction $\Sigma$ while also explicitly incorporating the section-wise feedback $\mathcal{F}$. This approach allows the model to maintain the original design intent while correcting localized issues such as spacing inconsistencies, text contrast problems, or missing visual elements.
The refiner operates under a controlled instruction template that parses feedback, preserves the global layout structure, and selectively edits only the affected regions.

Only sections with scores $\leq \tau$ (default $\tau=4$) are passed to the LLM for correction. This minimizes unnecessary regeneration and stabilizes layout consistency.
After refinement, the new HTML $\hat{H}'$ is re-rendered and re-evaluated using the same section-wise metrics, enabling quantitative comparison between pre- and post-refinement results.

Empirically, models learn to perform local edits rather than full regeneration when guided by structured feedback, often issuing natural corrections. This section-wise refinement loop closes the generation–evaluation gap by turning interpretive feedback into precise editing instructions.

\begin{figure*}[t]
\footnotesize
\centering
\begin{tcolorbox}[colback=white,colframe=blue!60,title=Refinement Prompt]
\textbf{System Prompt:}
\vspace{1mm}

You are a senior front-end engineer and UX designer with expertise in premium web design. Your task is to refine the provided HTML based on the original design instruction and a specific list of corrective tasks.

\vspace{2mm}
\textbf{CRITICAL RULES:}
\begin{enumerate}
    \item NEVER change image paths or filenames starting with './asset/' - these must remain EXACTLY as they are!
    \item REWRITE problematic sections to fix issues - don't just tweak, completely rewrite them.
    \item For severe issues (score 1-2), completely rewrite the affected sections.
    \item Maintain semantic HTML5 structure and accessibility standards.
    \item Use proper CSS Grid/Flexbox for layouts.
    \item Ensure responsive design with mobile-first approach.
\end{enumerate}

\textbf{PRIORITY ORDER:}
\begin{itemize}
    \item Image size/aspect ratio issues (MSA) - CRITICAL
    \item Alignment issues (ALN) - CRITICAL
    \item Spacing issues (SPC) - CRITICAL
    \item Media placement (MP) - HIGH
    \item Text readability (TR) - HIGH
    \item Other improvements - MEDIUM
\end{itemize}

\textbf{OUTPUT FORMAT:}
\begin{itemize}
    \item Return ONLY the complete, valid HTML document.
    \item No explanations, no markdown code blocks, no additional text.
    \item Include necessary comments within the HTML itself.
\end{itemize}

\hrulefill
\vspace{2mm}

\textbf{User Prompt:}
\vspace{1mm}

\textbf{Instruction}
\begin{tcolorbox}[colback=yellow!10!white, sharp corners, boxrule=0.5pt, title=This is the original instruction used for the initial generation.]
\{original\_instruction\}
\end{tcolorbox}

\textbf{Original HTML to Refine}
\begin{tcolorbox}[colback=gray!5!white, sharp corners, boxrule=0.5pt]
\{html\_content\}
\end{tcolorbox}

\textbf{Corrective Tasks}
\begin{tcolorbox}[colback=red!5!white, sharp corners, boxrule=0.5pt, title=These tasks are extracted from low-score evaluation feedback.]
1. \{task\_1\_from\_feedback\} \\
2. \{task\_2\_from\_feedback\} \\
...
\end{tcolorbox}

\textbf{Implementation Rules \& Critical Constraints}
\begin{itemize}
    \item \textbf{REAL IMAGE PATHS:} DO NOT change any image paths or filenames that start with './asset/' - keep them EXACTLY as they appear.
    \item \textbf{FEEDBACK PATHS:} Ignore paths like './assets/element\_XX\_parent-xxx\_class-xxx.png' - these are structural descriptions from feedback (not real files).
    \item \textbf{Semantic Structure \& Accessibility:} Use proper HTML5 elements, hierarchical headings, alt text for images, and ensure keyboard navigation.
    \item \textbf{Layout \& Responsive Design:} Use modern CSS (Grid/Flexbox) for a mobile-first, fluid layout. Maintain consistent spacing and alignment.
    \item \textbf{IMAGE PATHS CONSTRAINT:} DO NOT CHANGE ANY image src paths that start with './asset/' under any circumstances.
    \item \textbf{Output format:} Return ONLY the complete, valid HTML document without any additional text or markdown.
\end{itemize}
\end{tcolorbox}
\caption{The unified refinement prompt template used to generate the corrected HTML $\hat{H}'$. The prompt combines three key inputs: the original instruction ($\Sigma$), the initial HTML code to be modified ($\hat{H}$), and a list of corrective tasks. These tasks are derived by filtering the structured evaluation feedback ($\mathcal{F}$) to include only low-scoring items (e.g., score $\le 4$), effectively converting qualitative feedback into precise editing instructions for the model.}
\label{fig:refinement-prompt}
\end{figure*}

\section{Keyword List for Crawling}
\label{appendix:keywords}
To construct our dataset, we used domain-specific keyword lists to guide the crawling process. 
Table~\ref{tab:keywords} summarizes the main keyword categories and their coverage.

\begin{table*}[t]
\centering
\small
\caption{Keyword Categories for Crawling}
\begin{tabular}{p{3cm}p{12cm}}
\toprule
\textbf{Category} & \textbf{Example Keywords} \\
\midrule
SaaS &
crm, sales automation, contact management software, pipeline management, lead management software, customer support software, help desk software, live chat software, customer success platform, project management, task management, team collaboration software, workflow automation, time tracking software, resource management software\\
Finance and Insurance &
business loans, small business loans, mortgage refinancing, home equity loans, 
auto loans, loan consolidation, debt relief, credit repair, credit monitoring, 
robo advisor, wealth management, stock trading apps, options trading, forex trading, 
crypto trading platform, bitcoin exchange, nft marketplace, financial planning app, 
budgeting app, retirement accounts, ira accounts, 401k rollover, annuities, 
life insurance, term life insurance, whole life insurance, universal life insurance, 
health insurance, medicare plans, dental insurance, vision insurance, pet insurance, 
auto insurance, renters insurance, homeowners insurance, travel insurance, 
disability insurance, business insurance, liability insurance, cyber insurance, 
malpractice insurance, commercial auto insurance, umbrella insurance, 
insurance broker, insurance comparison, insurance quotes, insurance online, 
tax filing software, online bookkeeping, payroll services, invoice factoring, 
business bank account, checking account, savings account, online bank, 
neobank, high yield savings, cd accounts, money transfer app, remittance service, 
payment gateway, merchant account, point of sale system, cash advance loans, 
capital loans, invoice financing, kyc platform, aml compliance, fraud detection software, 
identity verification software, credit scoring platform, procurement financing, 
supply chain finance \\
Health and Wellnes &
telemedicine, telehealth app, online therapy, therapy near me, counseling online, 
mental health app, meditation app, mindfulness app, stress management app, 
addiction recovery program, rehab centers, nutrition coaching, diet app, 
calorie counter app, meal planner app, weight loss program, weight loss app, 
fitness app, workout tracker, gym membership, personal trainer app, 
home workout program, yoga classes online, pilates classes online, dance fitness app, 
running app, cycling app, smartwatch fitness tracker, supplements online, 
protein powder online, vitamins subscription, skincare products online, 
acne treatment online, hair loss treatment, erectile dysfunction treatment, 
birth control online, period tracker app, fertility app, pregnancy app, 
baby monitor app, pediatric care online, senior care services, hearing aids online, 
dental aligners online, invisalign alternative, braces online, teeth whitening kit, 
contact lenses online, glasses online, blue light glasses, vision correction surgery, 
cosmetic surgery clinic, plastic surgery online consultation, botox treatment online, 
laser hair removal booking, dermatology online, skincare telehealth, 
wellness retreat booking, spa booking, massage booking online, chiropractor near me, 
acupuncture booking, physiotherapy booking, allergy treatment, asthma treatment online, 
sleep tracker app, cbt app \\
E-commerce &
fashion sale, buy shoes online, sneaker release, handbag sale, jewelry online, 
watches online, wedding rings online, luxury bags online, fast fashion clothing, 
sustainable clothing brand, kids clothing online, maternity wear online, 
furniture sale, sofa online, mattress online, bed in a box, dining set sale, 
office chair sale, ergonomic chair, standing desk sale, gaming chair online, 
electronics sale, laptop deals, buy iphone online, refurbished phones, 
headphones sale, bluetooth speaker online, tv deals, smart home devices, 
smartwatch sale, camera sale, drone sale, 3d printer sale, 
kitchen appliances sale, blender online, air fryer sale, coffee machine deals, 
vacuum cleaner online, robot vacuum sale, washing machine deals, 
fridge sale, dishwasher online, oven sale, cookware set online, 
pet supplies subscription, pet food online, cat litter subscription, 
dog toys online, pet insurance online, meal kit delivery, food delivery app, 
grocery delivery, wine subscription, beer subscription, coffee subscription, 
subscription box service, beauty box subscription, skincare subscription, 
makeup online, fragrance online, nail polish online, sunscreen online, 
shampoo online, conditioner online, hair dye online, razor subscription, 
contact lenses subscription\\
Travel and Lifestyle &
cheap flights, flight booking, airline tickets online, last minute flights, 
business class deals, vacation packages, all inclusive resorts, 
honeymoon packages, family vacation deals, cruise booking, cruise deals, 
car rental deals, airport transfers, travel insurance online, travel credit cards, 
hotel booking, boutique hotels, luxury hotels, budget hotels, 
hostel booking, vacation rentals, airbnb alternatives, camping gear sale, 
glamping booking, rv rental, tour packages, local tours booking, 
city pass deals, theme park tickets, attraction tickets online, concert tickets online, 
festival tickets, sports tickets online, travel visa services, language learning app, 
study abroad programs, exchange student programs, international sim card, 
esim for travel, travel wifi rental, guided tours, hiking trips, adventure tours, 
ski packages, scuba diving trips, safari booking, yacht charter, private jet booking \\
Education and Professional Services &
online mba, business school online, coding bootcamp, data science bootcamp, 
ui ux bootcamp, product management course, leadership training, executive coaching, 
language learning online, english tutoring, math tutoring online, 
sat prep online, gmat prep online, gre prep online, ielts prep online, 
professional certification online, cybersecurity certification, 
cloud certification, aws certification training, microsoft certification training, 
google certification training, digital marketing course, seo course online, 
content writing course, copywriting course, design course online, 
video editing course, music lessons online, art classes online, photography course \\
\bottomrule
\end{tabular}
\label{tab:keywords}
\end{table*}

\end{document}